\documentclass[10pt,twocolumn,letterpaper]{article}

\usepackage{iccv}              

\usepackage{color}
\usepackage{xcolor}
\usepackage{colortbl}
\definecolor{Gray}{gray}{0.9}
\usepackage{threeparttable}
\usepackage{multirow}
\usepackage{array}
\usepackage{makecell}
\usepackage{bbding}
\usepackage{pifont}
\usepackage{tcolorbox}
\usepackage{amsmath}
\definecolor{iccvblue}{rgb}{0.21,0.49,0.74}
\definecolor{Gray}{gray}{0.9}
\definecolor{ours}{RGB}{244,237,252}
\definecolor{mygreen}{RGB}{0,128,0}  
\definecolor{myred}{RGB}{255,0,0}
\definecolor{ym_orange}{HTML}{D47C3A}

\usepackage[pagebackref,breaklinks,colorlinks,allcolors=iccvblue]{hyperref}

\def\paperID{xxxx} 
\def\confName{ICCV}
\def\confYear{2025}

\title{ViSpeak: Visual Instruction Feedback in Streaming Videos}

\author{%
Shenghao Fu$^{1,2,4,\dag}$, Qize Yang$^{2,\dag}$, Yuan-Ming Li$^{1,4}$, Yi-Xing Peng$^{1,2,4}$, Kun-Yu Lin$^{1,4}$, \\ Xihan Wei$^{2}$, Jian-Fang Hu$^{1,4}$,
{Xiaohua Xie$^{1,4,5}$\thanks{: Corresponding authors are Xiaohua Xie and Wei-Shi Zheng. $^{\dag}$: Equal Contribution. Work was done when Shenghao Fu and Yi-Xing Peng were interns at Alibaba.}, Wei-Shi Zheng$^{1,3,4,6*}$} \\
$^1$School of Computer Science and Engineering, Sun Yat-sen University, China;\\
$^2$Tongyi Lab, Alibaba Group; \quad $^3$Peng Cheng Laboratory, China; \\
$^4$Key Laboratory of Machine Intelligence and Advanced Computing, Ministry of Education, China;\\
$^5$Guangdong Province Key Laboratory of Information Security Technology, China; \\
$^6$Pazhou Laboratory (Huangpu), China\\
{\small fushh7@mail2.sysu.edu.cn, qize.yqz@alibaba-inc.com, xiexiaoh6@mail.sysu.edu.cn, wszheng@ieee.org} \\
{\small ViSpeak: \url{https://github.com/HumanMLLM/ViSpeak}} \\
{\small ViSpeak-Bench: \url{https://github.com/HumanMLLM/ViSpeak-Bench}}
}

\begin{document}
\maketitle

\begin{figure*}[t]
  \centering
  \vspace{-0.6em}
  \includegraphics[width=0.9\linewidth]{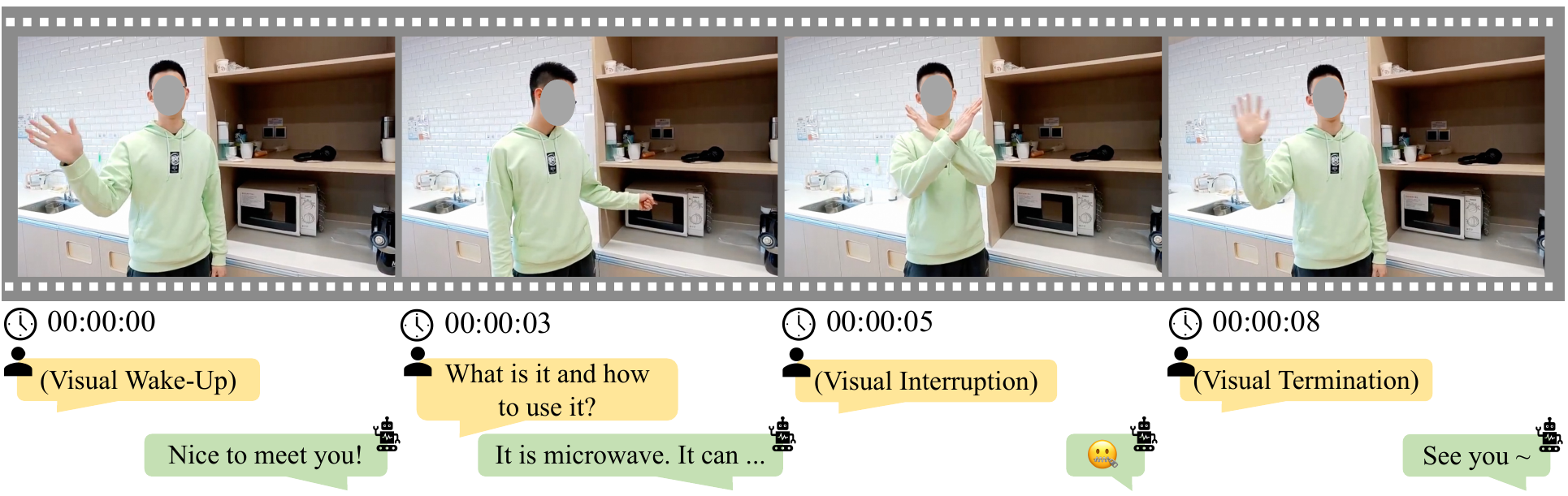}
  \vspace{-0.6em}
  \caption{Examples of some actions in Visual Instruction Feedback task, which are Visual Wake-Up, Visual Reference, Visual Interruption, and Visual Termination in order. The content in parentheses is displayed by body language instead of text or speech.}
  \vspace{-1.0em}
  \label{fig:task_example}
\end{figure*}

\begin{abstract}
Recent advances in Large Multi-modal Models (LMMs) are primarily focused on offline video understanding. Instead, streaming video understanding poses great challenges to recent models due to its time-sensitive, omni-modal and interactive characteristics. In this work, we aim to extend the streaming video understanding from a new perspective and propose a novel task named \textbf{Visual Instruction Feedback} in which models should be aware of visual contents and learn to extract instructions from them. For example, when users wave their hands to agents, agents should recognize the gesture and start conversations with welcome information. Thus, following instructions in visual modality greatly enhances user-agent interactions. To facilitate research, we define seven key subtasks highly relevant to visual modality and collect the \textbf{ViSpeak-Instruct} dataset for training and the \textbf{ViSpeak-Bench} for evaluation. Further, we propose the \textbf{ViSpeak} model, which is a SOTA streaming video understanding LMM with GPT-4o-level performance on various streaming video understanding benchmarks. After finetuning on our ViSpeak-Instruct dataset, ViSpeak is equipped with basic visual instruction feedback ability, serving as a solid baseline for future research.
\end{abstract}

\section{Introduction}

Recent Large Video Language Models~\cite{li2024llavaonevision, zhang2024llavanextvideo, zhang2024video, chen2025sharegpt4video, fu2025vita15, liu2025ola} excel at fine-grained spatial perception, long-term temporal reasoning, and comprehensive spatiotemporal understanding. In the offline setting, the entire video is provided and the complete context can be modeled. However, in streaming video understanding, models can not access the entire video. Video content continuously arrives, and the model must make decisions based on the information available so far while continuously processing incoming future data, which poses great challenges to recent LMMs.

Three key differences exist between streaming and offline video understanding: 
\textit{First, the question answers in streaming video understanding are time-sensitive.} The answers for the same question ``What is happening now?'' vary at different timestamps and the model should output the answer at a proper time. 
\textit{Second, steaming videos are always accompanied by streaming audios}, making problems as omni-modal ones. 
\textit{Third but most importantly, streaming video understanding is distinguished by its \textbf{interaction characteristic}.} The interaction characteristic encompasses three folds: \textbf{1) non-awakening interaction} where users can interact with agents at any time, \textbf{2) interruption} where users can stop the answer or change the topic at any time, and \textbf{3) proactive output} where agents can also express their mind at a proper time. Despite its significance, the interaction characteristic has \textit{been largely overlooked} by the community. MMDuet~\cite{mmduet} and Dispider~\cite{qian2025dispider} conducted preliminary explorations on proactive output to point out a specific event when it occurs based on user prompt. However, the prompts do not always exist, especially for an unintentional event or during communications. VITA~\cite{fu2024vita} uses dual models to decide when to respond to instructions in audio but it can not respond to visual contents.

In this work, we dive deeper into the interaction characteristic of streaming video understanding and introduce a new task named \textbf{Visual Instruction Feedback} to explore the instructions in the visual modality. We restrict the feedback primarily in conversational scenarios and define it as \textit{a kind of feedback towards visual contents to provide in-time interaction with users and necessary assistance effectively}. In this task, we select seven representative subtasks, including: 1) Visual Wake-Up: users use body language to start the conversation, 2) Anomaly Warning: agents provide in-time warnings or advice based on accidental events, 3) Gesture Understanding: agents respond to gestures from humans in conversations, 4) Visual Reference: users use body language to refer to a specific object, 5) Visual Interruption: users use body language to stop agents speaking, 6) Humor Reaction: agents share feedback to funny things with users, and 7) Visual Termination: users use body language to end the conversation. Examples are shown in \Cref{fig:task_example}.
To facilitate exploration, we collect the \textbf{ViSpeak-Bench} benchmark containing 1,000 videos and 1,000 QA pairs and the \textbf{ViSpeak-Instruct} training dataset containing 34k samples. As shown in \Cref{tab:benchmark_compare}, ViSpeak-Bench is the first comprehensive benchmark to evaluate the ability to respond to instructions in visual modality.

However, to the best of our knowledge, none of the open-sourced models can perform the Visual Instruction Feedback task even after finetuning on our dataset, especially for the visual interruption subtask, as they adopt a turn-taking chat template and the agent will fully express its mind without interruption before analyzing new user inputs.
Thus, we propose the \textbf{ViSpeak} model which is finetuned from an existing omni-model using a novel three-stage finetuning procedure. In the first template alignment stage, we adapt the offline model to a streaming input template while preserving the original offline capacities. The template supports taking the user's input and the model's responses as inputs at the same time, making the two input streams fully time-aligned. This template also supports interruption when the model is speaking. In the second streaming finetuning stage, we enhance the model's streaming question-answering ability and proactive output ability. The resulting model achieves SOTA performance on the StreamingBench~\cite{lin2024streamingbench} and OVO-Bench~\cite{li2025ovo}, achieving 62.00 and 61.08 overall scores, separately, which are comparable with GPT-4o. Finally, we finetune the model on our collected ViSpeak-Instruct dataset which serves as a solid baseline for the Visual Instruction Feedback task.

In summary, our contributions are three folds:
\begin{enumerate}
    \item We propose a novel streaming video understanding task named Visual Instruction Feedback, which requires the model to actively respond to visual contents. This task greatly enhances human-agent interactions. 
    \item To support exploration, we manually collect the ViSpeak-Bench benchmark and the ViSpeak-Instruct training dataset. We also provide some analysis based on the evaluation results of existing models.
    \item We also propose a strong baseline ViSpeak for the new task, which is finetuned from an existing omni-modal model with three-stage finetuning. ViSpeak not only preserves offline understanding capacities but also achieves SOTA performance on streaming video understanding benchmarks.
\end{enumerate}

\section{Related Work}
\subsection{Large Multi-Modal Model}

Recent Large Multi-modal Models (LMMs) have rapidly evolved from image understanding models~\cite{llava, liu2024improved, chen2024sharegpt4v, wang2024pargo} to video understanding models~\cite{zhang2024llavanextvideo, li2024llavaonevision, chen2025sharegpt4video, zhang2024video} and even to omni-modal understanding models~\cite{fu2024vita, fu2025vita15, li2025baichuan, liu2025ola, zhao2025humanomni}. With high-quality instruction turning data~\cite{chen2024sharegpt4v, chen2025sharegpt4video, wang2024tarsier}, improved training recipes~\cite{li2024llavaonevision, wang2024tarsier}, and well-designed model architecture~\cite{wang2024pargo, xu2024slowfast, liu2024oryx, li2024techcoach}, recent LMMs achieve fine-grained multi-modal alignment and extend their abilities of comprehensive image-level understanding~\cite{wang2024qwen2, li2024llavaonevision}, fine-grained region perception~\cite{lai2024lisa,asv2,jiang2024chatrex, fu2025llmdet}, long-term temporal reasoning~\cite{qian2025streaming, zhang2024internlm, zhang2024omnilive}, timestamp awareness~\cite{liu2024bench, wang2024grounded} and even human mind or emotional understanding~\cite{hyun2023smile,xie2024funqa,zhao2025humanomni, yang2025omni}. Although great progress has been made, recent video LMMs are primarily focused on offline videos where the entire video is provided for understanding.

\subsection{Streaming Video Understanding}

In practical human-agent interactions, LMMs should process streaming videos, which has drawn great attention in recent years. Many streaming video understanding benchmarks~\cite{lin2024streamingbench, li2025ovo, xiong2025streaming, yang2025svbench} have been proposed, which have simultaneously spurred the development of many streaming video LMMs. VideoLLM-online~\cite{chen2024videollm}, as a pioneer model, proposes a LIVE framework to process streaming videos, which uses Streaming Loss to learn when to speak. Subsequently, different models focus on different challenges in streaming video understanding. Flash-VStream~\cite{zhang2024flash} and IXC2.5-OL~\cite{zhang2024omnilive} propose some memory mechanisms to handle long context in streaming videos. Dispider~\cite{qian2025dispider} and MMDuet~\cite{mmduet} focus on proactive output, the former disentangling perception, decision, and reaction while the latter introduces two additional heads.
Mini-Omni2~\cite{xie2024miniomni2} and VITA 1.5~\cite{fu2025vita15} pay more attention to end-to-end real-time speech interaction. STREAMCHAT~\cite{xiong2025streaming} and StreamingChat~\cite{yang2025svbench} aim to tackle the multi-round conversation problem. In this work, we extend the streaming video understanding problem from a new perspective and introduce a new Visual Instruction Feedback task.

\section{Visual Instruction Feedback Task}

\begin{table}[t]
  \centering
  \vspace{-0.6em}
  \resizebox{\linewidth}{!}{
      \begin{tabular}{lccccccc}
        \hline
        Benchmark & \#Videos & \#QA Pairs & Time & Streaming & PO & Visual Instruct & Anno \\
        \hline
        ActivityNet-QA~\cite{yu2019activitynetqa} & 800 & 8,000 & \textcolor{myred}{\ding{55}} & \textcolor{myred}{\ding{55}} & \textcolor{myred}{\ding{55}} & \textcolor{myred}{\ding{55}} & Manual \\
        NExT-QA~\cite{xiao2021next} & 1,000 & 8,564 & \textcolor{myred}{\ding{55}} & \textcolor{myred}{\ding{55}} & \textcolor{myred}{\ding{55}} & \textcolor{myred}{\ding{55}} & Auto \\
        MVBench~\cite{li2024mvbench} & 3,641 & 4,000 & \textcolor{myred}{\ding{55}} & \textcolor{myred}{\ding{55}} & \textcolor{myred}{\ding{55}} & \textcolor{myred}{\ding{55}} & Auto \\
        Video-MME~\cite{fu2024videomme} & 900 & 2,700 & \textcolor{myred}{\ding{55}} & \textcolor{myred}{\ding{55}} & \textcolor{myred}{\ding{55}} & \textcolor{myred}{\ding{55}} & Manual \\
        ET-Bench~\cite{liu2024bench} & 7,002 & 7,289 & \textcolor{mygreen}{\checkmark} & \textcolor{myred}{\ding{55}} & \textcolor{myred}{\ding{55}} & \textcolor{myred}{\ding{55}} & Manual \\
        StreamingBench~\cite{lin2024streamingbench} & 900 & 4,500 & \textcolor{mygreen}{\checkmark} & \textcolor{mygreen}{\checkmark} & \textcolor{mygreen}{\checkmark} & \textcolor{myred}{\ding{55}} & Mixed \\
        \rowcolor{ours} ViSpeak-Bench & 1,000 & 1,000 & \textcolor{mygreen}{\checkmark} & \textcolor{mygreen}{\checkmark} & \textcolor{mygreen}{\checkmark} & \textcolor{mygreen}{\checkmark} & Mixed \\
        \hline
      \end{tabular}
    }
  \vspace{-0.6em}
  \caption{Comparison between ViSpeak-Bench and other video benchmarks. `Time' means the dataset is time-sensitive. `PO' denotes the dataset to evaluate the proactive output ability.}
  \vspace{-0.6em}
  \label{tab:benchmark_compare}
\end{table}

\subsection{Task Definition}

In this work, we define a new task named \textbf{Visual Instruction Feedback} for streaming video understanding. Formally, we define the feedback as \textit{a kind of feedback towards visual contents to provide in-time interaction with users and necessary assistance effectively}. We also restrict the feedback primarily to conversational scenarios. In this task, users may not provide explicit instructions in text or audio format. The agent should analyze visual inputs and express its mind accordingly. Assume a video stream $X_{[0,+\infty)}$ with infinite length. An action $A_{[t_1, t_2]}$ or an event $E_{[t_1, t_2]}$ may appear at any time $[t_1, t_2]$ and the model should recognize them and provide feedback within a limited time span $[t_1, t_2+T]$.
As the task focuses on conversational scenarios, the agent should provide responses from a second-person perspective.

According to the definition, we summarize seven key subtasks. Before conversations:
\begin{enumerate}
    \item \textbf{Visual Wake-Up (VW)}. Unlike keyword-based (like Siri) or VAD-based~\cite{fu2024vita, zhang2024omnilive} wake-up, visual wake-up needs the model response to salutations from users.
    \item \textbf{Anomaly Warning (AW)}. In this task, models need to identify accidental events (\eg fighting, explosion) or unintentional actions (\eg falling down) and provide in-time warnings, advice, or help.
\end{enumerate}
During conversations:
\begin{enumerate}[start=3]
    \item \textbf{Gesture Understanding (GU)}. Gestures play a vital role in conversations, even serving as a short response from users, like ``OK'', ``GOOD'', ``ONE'', ``TWO''. Models should understand human gestures and provide the corresponding feedback.
    
    \item \textbf{Visual Reference (VR)}. In many cases, it is difficult to describe an object or where the object is precisely, but it can be done by pointing it out with the fingers, such as ``What is this''. Models should identify which object is referenced and answer questions from uses.
    
    \item \textbf{Visual Interruption (VI)}. When users are not satisfied with the model's response or want to change the topic, they may interrupt the model with some body language, like the stop gesture. Models should stop generating the remaining responses when receiving these signals.

    \item \textbf{Humor Reaction (HR)}. Humor understanding is one of the key abilities of humans. Reacting properly to funny things provides necessary emotional value to users.

    \item \textbf{Visual Termination (VT)}. Visual termination is the action to end conversations. Although the actions in wake-up and termination may be the same (\ie wave hands), they can be classified by the context where the action at the beginning of conversations is visual wake-up, otherwise visual termination. Models should be aware of contexts and start or end conversations properly.
    
\end{enumerate}
Although there are many other scenarios where agents should talk to users actively, for example, sign language (we exclude it due to the technical complexity and its variability across the world), we believe the subtasks above cover common scenarios in daily life. Some examples are shown in \Cref{fig:task_example}. More visualizations are shown in Supplement.

\begin{table}[t]
  \centering
  \vspace{-0.6em}
  \resizebox{0.9\linewidth}{!}{
      \begin{tabular}{lccc}
        \hline
        Subtask & \#Videos & \#QA Pairs & QA Type \\
        \hline
        Visual Wake-Up & 100 & 100 & Open-Ended \\
        Anomaly Warning & 200 & 200 & Open-Ended \\
        Gesture Understanding & 200 & 200 & Open-Ended \\
        Visual Reference & 200 & 200 & Multi-Choice \\
        Visual Interruption & 100 & 100 & Open-Ended \\
        Humor Reaction & 100 & 100 & Open-Ended \\
        Visual Termination & 100 & 100 & Open-Ended \\
        \hline
        \rowcolor{ours} ViSpeak-Bench & 1,000 & 1,000 & \\
        \hline
      \end{tabular}
    }
  \vspace{-0.6em}
  \caption{ViSpeak-Bench benchmark statistics. ViSpeak-Bench contains 7 subtasks with 1,000 videos and 1,000 QA pairs.}
  \vspace{-0.6em}
  \label{tab:benchamrk_stats}
\end{table}

\subsection{Dataset Construction}

\noindent{\textbf{Video Collection and Annotation.}} We collect videos from both open-sourced datasets and our self-collected datasets.

For open-sourced datasets, we use anomaly videos in Holmes-VAU~\cite{zhang2024holmes} and unintentional videos in OOPS~\cite{epstein2020oops} for Anomaly Warning and HumorQA in FunQA~\cite{xie2024funqa} for Humor Reaction. All the datasets above are annotated with timestamps and event descriptions. We simply use GPT-4o to rewrite the annotations in a conversational tone to simulate conversations. For Gesture Understanding, we select 10 common gestures from Jester~\cite{materzynska2019jester}, each with 400 videos.

For other subtasks, we manually record the videos by ourselves. To ensure diversity, we recruit a team of 610 people (346 men and 264 women) with an age ranging from 10 to 70 years old from 5 provinces. For each kind of subtask, we carefully designed diverse conversation scripts and instructed participants to follow these scripts during filming to simulate human-computer interaction scenarios. Videos are recorded in various environments, including homes, offices, factories, warehouses, supermarkets, wild, and many others. In summary, we collect 1,185 videos for Visual Interruption, 4,689 videos for Visual Reference, 1,188 videos for Visual Wake-Up and Termination, and 1,507 videos for Gesture Understanding. For each video, we manually annotate the accurate timestamps for each body language, making the videos suitable for streaming video understanding. The corresponding scripts for each video are used as the annotations. For the Gesture Understanding subtask, in addition to the 10 gestures in Jester~\cite{materzynska2019jester}, we further add 10 gestures commonly used during conversations, with a total of 20 gestures. We also design 5 gestures for Visual Interruption. More details can be found in Supplement. 

\vspace{0.2em}\noindent{\textbf{Dataset Enhancement.}} Although we have made great efforts to make the dataset large and diverse, the self-collected data still cannot cover infinite scenarios in the real world, especially for gesture understanding. To alleviate the problem, we augment the dataset with some offline video understanding datasets. The motivation is that the model can improve its social intelligence from the perspective of a bystander, simulating a child, who observes the conversations between adults. The offline data have no timestamp annotations and questions are appended at the end of the video.

Specifically, we select SMILE~\cite{hyun2023smile} for understanding funny things and IntentQA~\cite{li2023intentqa} and Social-IQ~\cite{zadeh2019social} for learning body languages. Further, we manually review the videos in Social-IQ~\cite{zadeh2019social} dataset and re-annotate some common body languages lasting longer than 1 second in conversations. For each action, we point out what action it is and why the speaker did the action in the context, which helps the model study the meaning of common body language in the wild. In summary, 678 videos with 1,861 annotations are collected. Examples can be found in Supplement.

\vspace{0.2em}\noindent{\textbf{Quality Verification.}} While recording videos, we will provide guidelines to participants to ensure the quality of the raw videos. The videos are then sent for human annotation. Low-quality data will be rejected and re-recorded. Some well-performed annotators will conduct spot checks on the annotation results. If significant quality issues are identified during annotation, the corresponding annotator will undergo retraining, and the data will be re-annotated.

\begin{table}[t]
  \centering
  \vspace{-0.6em}
  \resizebox{\linewidth}{!}{
      \begin{tabular}{lccccccc}
        \hline
        Subtask & Data Source & Data Type & \#Samples & Ratio \\
        \hline
        Visual Wake-Up & self-collected data & online & 1k & 0.03 \\
        \hline
        \multirow{2}{*}{Anomaly Warning} & OOPS~\cite{epstein2020oops} & online & 3k & 0.09 \\
        & HIVAU~\cite{zhang2024holmes} & online & 3k & 0.09 \\
        \hline
        \multirow{6}{*}{Gesture Understanding} & Jester~\cite{materzynska2019jester} & online & 4k & 0.12 \\
        & self-collected data & online & 4k & 0.12 \\
        & Social-IQ~\cite{zadeh2019social} & offline & 2k & 0.06 \\
        & IntentQA~\cite{li2023intentqa} & offline & 5k & 0.15 \\
        & SocialIQA~\cite{sap2019socialiqa} & offline & 0.5k & 0.02 \\
        & self-collected data & offline & 1k & 0.03 \\
        \hline
        Visual Reference & self-collected data & online & 5k & 0.15 \\
        \hline
        Visual Interruption & self-collected data & online & 1k & 0.03 \\
        \hline
        \multirow{2}{*}{Humor Reaction} & FunQA~\cite{xie2024funqa} & online & 2k & 0.06 \\
        & SMILE~\cite{hyun2023smile} & offline & 1k & 0.03 \\
        \hline
        Visual Termination & self-collected data & online & 1k & 0.03 \\
        \hline
        \rowcolor{ours} ViSpeak-Instruct & & & 34k & 1 \\
        \hline
      \end{tabular}
    }
    \vspace{-0.6em}
  \caption{Task and sample distribution in ViSpeak-Instruct.}
  \vspace{-0.6em}
  \label{tab:training_dataset_stats}
\end{table}

\begin{figure*}[t]
  \centering
  \includegraphics[width=0.9\linewidth]{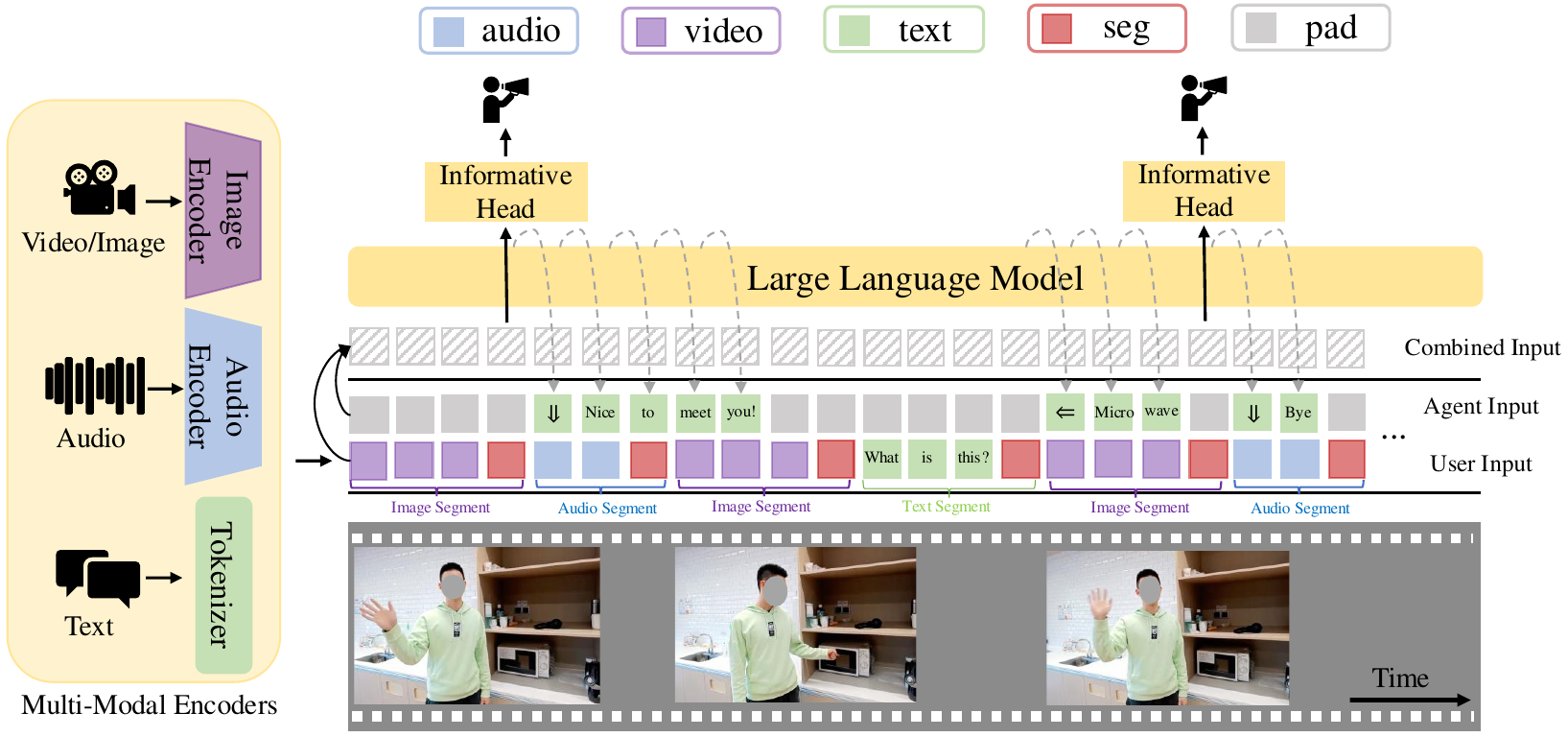}
  \vspace{-0.6em}
  \caption{ViSpeak is an omni-modality LMM with multiple encoders and a LLM. To support streaming video analysis, ViSpeak takes two input streams as inputs, one for user inputs and one for self-generated outputs. Two streams will be combined into a single one before sending to LLM. An informative head is trained for visual proactive output.}
  \vspace{-0.6em}
  \label{fig:model_structure}
\end{figure*}

\vspace{0.2em}\noindent{\textbf{Data Partition and Dataset Statistics.}} With the collected data above, we manually select some representative videos to construct the ViSpeak-Bench evaluation dataset. For Visual Wake-Up, Visual Termination and Visual Interruption, we select 100 videos for each subtask with actions lasting 2 seconds. For Visual Reference, we carefully select 200 videos with multiple objects. The referenced object may appear at any location within the frame and is not necessarily positioned at the center. We formulate this subtask as a multi-choice problem during evaluation and manually annotate each video with three other confusing options which are also displayed in the video, ensuring the answer is highly related to visual reference. For Humor Reaction, we select 100 humorous videos in FunQA that are only relevant to visual content. For Gesture Understanding and Anomaly Warning, we randomly select 200 videos for testing. The remaining data are contained in ViSpeak-Instruct and used for training. The statistics of ViSpeak-Bench and ViSpeak-Instruct are summarized in \Cref{tab:benchamrk_stats} and \Cref{tab:training_dataset_stats}.

\vspace{0.2em}\noindent{\textbf{Evaluation Metrics.}} 
In our task, we evaluate both the timing accuracy $\mathcal{T}_\text{acc}$ of the model's feedback (\ie, Time Accuracy) and the quality score $\mathcal{S}$ of its response text (\ie, Text Score), and then derive an overall score $\mathcal{O}$. 
For Time Accuracy, the model should response within the ground-truth time span $[t_1, t_2+T]$, where $T$ is the time margin we set. For subtask ${s}$, the accuracy of $T_\text{res}$ is measured based on whether it falls within this time window.
\begin{equation}
 \mathcal{T}^{s}_\text{acc} = \frac{1}{N^s}\sum_{i=1}^{N^s} \mathbb{I}\big(T_{\text{res}}^{(i)} \in [t_1, t_2+T]\big),
 \end{equation}
 where $N^{s}$ is the number of questions in each subtask.

For Text Score, the output of the model should accurately reflect the actions or events within the video and be consistent with historical context. The response must be positive and supportive, providing assistance to the user when necessary. Thus, we use the dialogue history and ground truth as references, designing different prompts for different subtasks. We use GPT-4o as the judge model, scoring the responses on a scale from 0 to 5 (see Supplement for more details). Note that, for the visual reference task, we use multi-choice questions for evaluation and rescale the score for this task to range from [0,5], while its response time accuracy is always set to 1. 

Finally, the overall score is calculated as:
\begin{equation}
 \mathcal{O} = \frac{1}{N}\sum_{s=1}^{N} \mathcal{T}^{s}_\text{acc}\times \mathcal{S}^{s},
 \end{equation}
where $N$ is the number of subtasks.

\section{The ViSpeak Model}

\subsection{Model Architecture}

In order to accomplish the Visual Instruction Feedback task, we design a model named \textbf{ViSpeak} as shown in \Cref{fig:model_structure}, which is an omni-model with an image encoder to extract image or video features, an audio encoder for encoding both audios and music, and a large language model to integrate multimodal features and conduct analyses to fulfill the relevant instructions. However, the turn-taking chat template with explicit role control is not suitable for interruption. Inspired by Moshi~\cite{defossez2024moshi}, we design a \textbf{two-stream chat template}, one for user inputs and one for agent historical outputs, so that the model can continuously process user inputs while outputting the next tokens. Thus, the model can adjust outputs based on upcoming inputs. Two input streams are combined into a single one before sending to the LLM. By default, we use a linear layer to predict the weights for a weighted sum of the two streams.
In the ablation study part, we have tried many kinds of combination methods and found that they perform similarly.

Further, we segment the streaming inputs from users into multiple fragments, such as extracting one frame per second from the video and dividing the audio into 1-second snippets, subsequently organizing these segments in chronological order. Each segment is appended with a special $<$seg$>$ token and the LLM can only start to speak from it. To differentiate the response towards different kinds of instructions, answers for text, audio, and visual instructions start with ``$\Leftarrow$'', ``$\Rightarrow$'', and ``$\Downarrow$'' separately, following VITA~\cite{fu2024vita}. For Visual Interruption, the model can simply output ``$\Downarrow$ Stop!'' at a $<$seg$>$ token to stop generation.

When to speak is a key problem in streaming processing. We find that using the original language model head (next token prediction) is sufficient to handle text-answering problems, \ie the model can always output a ``$\Leftarrow$'' token at the end of a text segment. However, visual proactive output is a more challenging task and next token prediction can not manage the turn-taking problem well. Thus, we train another informative head to predict when to speak following MMDuet~\cite{mmduet}, which is a binary classification head to predict speaking or not. When the prediction score is above a predefined threshold, the model will respond to a visual instruction.
In this work, we do not take the turn-taking problem for audio modality into account for simplicity.

\begin{table*}[t]
  \centering
  \vspace{-0.6em}
  \resizebox{\linewidth}{!}{
      \begin{tabular}{lccc|ccccccccccc|ccccc|ccccc|c}
        \hline
        \multirow{2}{*}{Method} & \multirow{2}{*}{Params} & \multirow{2}{*}{Frames} & \multirow{2}{*}{Omni} & \multicolumn{11}{c|}{Real-Time Visual Understanding}  & \multicolumn{5}{c|}{Omni-Source Understanding} & \multicolumn{5}{c|}{Contextual Understanding} & \multirow{2}{*}{\textbf{Overall}} \\
        & & & & OP & CR & CS & ATP & EU & TR & PR & SU & ACP & CT & \textbf{All} & ER & SCU & SD & MA & \textbf{All} & ACU & MCU & SQA & PO & \textbf{All} &  \\
        \hline
        \rowcolor{Gray} \multicolumn{26}{c}{\textbf{Proprietary MLLMs}}\\
        \hline
        Gemini 1.5 pro~\cite{team2024gemini} & - & - & \checkmark & 83.43 & 77.94 & 89.24 & 81.65 & 79.17 & 83.92 & 83.93 & 60.32 & 74.87 & 49.22 & 77.39 & 52.40 & 50.80 & 80.40 & 87.60 & 67.80 & 52.80 & 42.40 & 59.20 & 45.10 & 51.06 & 70.26 \\
        GPT-4o~\cite{hurst2024gpt4o} & - & - & \checkmark & 80.66 & 76.98 & 86.67 & 73.81 & 75.95 & 85.48 & 75.00 & 70.66 & 65.99 & 43.09 & 74.54 & 53.60 & 32.40 & 49.00 & 68.80 & 50.95 & 50.40 & 42.80 & 52.40 & 56.86 & 49.06 & 64.31 \\
        Claude-3.5-sonnet & - & - & \ding{55} & 82.45 & 73.77 & 82.43 & 82.40 & 76.39 & 85.56 & 61.68 & 60.73 & 67.88 & 47.62 & 74.04 & 39.60 & 35.60 & 34.40 & 56.00 & 41.40 & 36.00 & 43.20 & 34.80 & 64.71 & 39.70 & 60.06 \\
        \hline
        \rowcolor{Gray} \multicolumn{26}{c}{\textbf{Open-Source Video MLLMs}}\\
        \hline
        LLaVA-OneVision~\cite{li2024llavaonevision} & 7B & 32 & \ding{55} & 82.83 & 77.34 & 83.23 & 83.33 & 72.05 & 74.77 & 73.15 & 68.29 & 71.10 & 41.97 & 74.27 & 41.20 & 26.10 & 43.20 & 52.80 & 40.83 & 35.08 & 30.40 & 30.00 & 29.55 & 31.68 & 58.56 \\
        MiniCPM-V~\cite{yao2024minicpm} & 8B & 32 & \ding{55} & 78.20 & 71.88 & 84.18 & 83.99 & 75.16 & 75.39 & 72.22 & 56.50 & 67.14 & 47.15 & 72.43 & 42.00 & 27.71 & 40.40 & 50.80 & 40.23 & 37.50 & 27.20 & 40.00 & 22.22 & 34.09 & 57.80 \\
        InternVL-V2~\cite{chen2024far} & 8B & 16 & \ding{55} & 73.84 & 65.63 & 78.80 & 82.03 & 71.43 & 72.90 & 73.15 & 63.01 & 65.44 & 42.49 & 70.11 & 44.80 & 28.11 & 47.20 & 50.80 & 42.73 & 35.08 & 27.20 & 42.80 & 40.91 & 35.40 & 57.28 \\
        Qwen2-VL~\cite{wang2024qwen2} & 7B & 1 fps & \ding{55} & 75.75 & 79.69 & 76.58 & 79.08 & 74.53 & 75.08 & 74.07 & 65.85 & 65.16 & 41.97 & 71.15 & 40.80 & 25.30 & 41.20 & 55.60 & 40.73 & 34.27 & 26.40 & 44.40 & 22.73 & 34.24 & 57.20 \\
        LLaVA-Next-Video~\cite{zhang2024llavanextvideo} & 32B & 64 & \ding{55} & 80.11 & 71.09 & 80.70 & 80.72 & 71.43 & 73.21 & 62.96 & 59.35 & 63.17 & 36.79 & 69.83 & 41.60 & 24.50 & 44.40 & 56.40 & 41.73 & 34.27 & 28.80 & 44.00 & 18.18 & 34.58 & 56.73 \\
        Video-LLaMA2~\cite{cheng2024videollama} & 7B & 32 & \checkmark & 59.95 & 60.16 & 62.97 & 60.46 & 54.66 & 46.11 & 41.67 & 46.75 & 48.16 & 34.72 & 52.58 & 43.60 & 23.29 & 35.20 & 41.60 & 35.92 & 28.23 & 26.00 & 21.20 & 0.00 & 23.54 & 43.30 \\
        Ola~\cite{liu2025ola} & 7B & 64 & \checkmark & 61.58 & 71.09 & 67.19 & 62.09 & 62.73 & 51.71 & 60.19 & 52.03 & 53.82 & 17.62 & 56.16 & 40.80 & 27.20 & 23.60 & 43.20 & 33.70 & 30.40 & 22.80 & 31.20 & 11.20 & 23.90 & 44.00 \\
        VITA 1.5~\cite{fu2025vita15} & 7B & 16 & \checkmark & 74.11 & 78.13 & 80.76 & 77.12 & 73.91 & 64.17 & 66.67 & 58.54 & 66.57 & 33.68 & 68.20 & 44.00 & 26.80 & 42.80 & 56.80 & 42.60 & 31.60 & 32.80 & 36.40 & 23.60 & 31.10 & 54.27 \\
        \hline
        \rowcolor{Gray} \multicolumn{26}{c}{\textbf{Open-Source Streaming MLLMs}}\\
        Flash-VStream~\cite{zhang2024flash} & 7B & - & \ding{55} & 25.89 & 43.57 & 24.91 & 23.87 & 27.33 & 13.08 & 18.52 & 25.20 & 23.87 & 48.70 & 23.23 & 25.91 & 24.90 & 25.60 & 28.40 & 26.00 & 24.80 & 25.20 & 26.80 & 1.96 & 24.12 & 24.04 \\
        VideoLLM-online~\cite{chen2024videollm} & 8B & 2 fps & \ding{55} & 39.07 & 40.06 & 34.49 & 31.05 & 45.96 & 32.40 & 31.48 & 34.16 & 42.49 & 27.89 & 35.99 & 31.20 & 26.51 & 24.10 & 32.00 & 28.45 & 24.19 & 29.20 & 30.80 & 3.92 & 26.55 & 32.48 \\
        IXC2.5-OL~\cite{zhang2024omnilive} & 7B & 64 & \ding{55} & 82.83 & 73.77 & 78.66 & 82.95 & 72.50 & 76.01 & 61.11 & 60.67 & 71.59 & 58.85 & 73.79 & - & - & - & - & - & - & - & - & - & - & - \\
        Dispider~\cite{qian2025dispider} & 7B & 1 fps & \ding{55} & 74.92 & 75.53 & 74.10 & 73.08 & 74.44 & 59.92 & 76.14 & 62.91 & 62.16 & 45.80 & 67.63 & 35.46 & 25.26 & 38.57 & 43.34 & 35.66 & 39.62 & 27.65 & 34.80 & 25.34 & 33.61 & 53.12 \\
        \rowcolor{ours} ViSpeak (Ours, s2) & 7B & 1 fps & \checkmark & 79.84 & 88.28 & 83.28 & 81.05 & 76.40 & 75.08 & 70.37 & 65.85 & 77.34 & 34.20 & 74.36 & 42.80 & 35.20 & 61.20 & 74.80 & 53.50 & 38.80 & 36.80 & 44.00 & 38.80 & 39.60 & 62.00 \\
        \rowcolor{ours} ViSpeak (Ours, s3) & 7B & 1 fps & \checkmark & 79.84 & 71.09 & 81.39 & 78.76 & 74.53 & 70.09 & 63.89 & 64.23 & 71.39 & 27.98 & 70.44 & 47.20 & 56.40 & 61.60 & 81.20 & 61.60 & 49.20 & 36.40 & 39.20 & 50.80 & 43.90 & 62.58 \\
        \hline
      \end{tabular}
    }
    \vspace{-0.6em}
  \caption{Performance on StreamingBench~\cite{lin2024streamingbench}. Results for ViSpeak trained after the second and third stage are reported.}
  \label{tab:streamingbench_result}
\end{table*}

\begin{table*}[t]
  \centering
  \vspace{-0.6em}
  \resizebox{\linewidth}{!}{
      \begin{tabular}{lcc|ccccccc|cccc|cccc|c}
        \hline
        \multirow{2}{*}{Method} & \multirow{2}{*}{Params} & \multirow{2}{*}{Frames} & \multicolumn{7}{c|}{Real-Time Visual Perception}  & \multicolumn{4}{c|}{Backward Tracing} & \multicolumn{4}{c|}{Forward Active Responding} & \multirow{2}{*}{\textbf{Overall}} \\
        & & & OCR & ACR & ATR & STU & FPD & OJR & \textbf{Avg.} & EPM & ASI & HLD & \textbf{Avg.} & REC & SSR & CRR & \textbf{Avg.} &  \\
        \hline
        \rowcolor{Gray} \multicolumn{19}{c}{\textbf{Proprietary MLLMs}}\\
        \hline
        Gemini 1.5 pro~\cite{team2024gemini} & - & - & 87.25 & 66.97 & 80.17 & 54.49 & 68.32 & 67.39 & 70.77 & 68.59 & 75.68 & 52.69 & 62.32 & 35.53 & 74.24 & 61.67 & 57.15 & 65.25 \\
        GPT-4o~\cite{hurst2024gpt4o} & - & - & 69.13 & 65.14 & 65.52 & 50.00 & 68.32 & 63.68 & 63.63 & 49.83 & 70.95 & 55.38 & 58.72 & 27.58 & 73.21 & 59.40 & 53.40 & 58.58 \\
        \hline
        \rowcolor{Gray} \multicolumn{19}{c}{\textbf{Open-Source Video MLLMs}}\\
        \hline
        Qwen2-VL~\cite{wang2024qwen2} & 72B & 64 & 72.48 & 56.88 & 77.59 & 52.25 & 74.26 & 61.41 & 65.81 & 51.52 & 73.65 & 63.44 & 62.87 & 37.68 & 60.10 & 45.00 & 47.59 & 58.76 \\
        Qwen2-VL~\cite{wang2024qwen2} & 7B & 64 & 69.13 & 53.21 & 63.79 & 50.56 & 66.34 & 60.87 & 60.65 & 44.44 & 66.89 & 34.41 & 48.58 & 30.09 & 65.66 & 50.83 & 48.86 & 52.70 \\
        LLaVA-Next-Video~\cite{zhang2024llavanextvideo} & 7B & 64 & 69.80 & 59.63 & 66.38 & 50.56 & 72.28 & 61.41 & 63.34 & 51.18 & 64.19 & 9.68 & 41.68 & 34.10 & 67.57 & 60.83 & 54.17 & 53.06 \\
        LLaVA-OneVision~\cite{li2024llavaonevision} & 7B & 64 & 67.11 & 58.72 & 69.83 & 49.44 & 71.29 & 60.33 & 62.79 & 52.53 & 58.78 & 23.66 & 44.99 & 24.79 & 66.93 & 60.83 & 50.85 & 52.88 \\
        InternVL-V2~\cite{chen2024far} & 8B & 16 & 68.46 & 58.72 & 68.97 & 44.94 & 67.33 & 55.98 & 60.73 & 43.10 & 61.49 & 27.41 & 44.00 & 25.79 & 57.55 & 52.92 & 45.42 & 50.05 \\
        LongVU~\cite{shen2024longvu} & 7B & 1 fps & 55.70 & 49.54 & 59.48 & 48.31 & 68.32 & 63.04 & 57.40 & 43.10 & 66.22 & 9.14 & 39.49 & 16.62 & 69.00 & 60.00 & 48.54 & 48.48 \\
        VITA 1.5~\cite{fu2025vita15} & 7B & 16 & 74.50 & 60.55 & 70.69 & 53.37 & 63.37 & 58.70 & 63.53 & 46.13 & 54.05 & 24.19 & 41.46 & 37.54 & 60.73 & 62.08 & 53.45 & 55.49 \\
        \hline
        \rowcolor{Gray} \multicolumn{19}{c}{\textbf{Open-Source Streaming MLLMs}}\\
        Flash-VStream~\cite{zhang2024flash} & 7B & 1 fps & 25.50 & 32.11 & 29.31 & 33.71 & 29.70 & 28.80 & 29.86 & 36.36 & 33.78 & 5.91 & 25.35 & 5.44 & 67.25 & 60.00 & 44.23 & 33.15 \\
        VideoLLM-online~\cite{chen2024videollm} & 8B & 2 fps & 8.05 & 23.85 & 12.07 & 14.04 & 45.54 & 21.20 & 20.79 & 22.22 & 18.80 & 12.18 & 17.73 & - & - & - & - & - \\
        \rowcolor{ours} ViSpeak (Ours, s2) & 7B & 1 fps & 75.17 & 58.72 & 71.55 & 51.12 & 74.26 & 66.85 & 66.28 & 59.93 & 48.65 & 63.98 & 57.52 & 33.81 & 68.52 & 60.42 & 54.25 & 61.08 \\
        \hline
      \end{tabular}
    }
    \vspace{-0.6em}
  \caption{Performance of various MLLMs on OVO-Bench~\cite{li2025ovo}. Results for ViSpeak trained after the second stage are reported.}
  \label{tab:ovobench_result}
\end{table*}

\subsection{A Three-Stage Finetuning Recipe}

Directly training a strong streaming model from scratch is resource-demanding. Thus we begin with a well-pretrained omni-modal offline model~\cite{fu2025vita15} and adopt a three-stage finetuning recipe to train the model.

In the first template alignment stage, we adapt the offline model to our streaming input template with the goal of not compensating for its offline multi-modal understanding ability. In this stage, we select 300k text data from Magie~\cite{xu2024magpie}, 665k image data from ShareGPT4V~\cite{chen2024sharegpt4v}, 1,335k video data from LLaVA-Video~\cite{zhang2024video}, 410k audio data from LibriSpeech~\cite{panayotov2015librispeech} and WavCaps~\cite{mei2024wavcaps}, and 121k cross-modality data from Ola~\cite{liu2025ola}, with a total of 2.7M data for training. To save computation, we compress the data by concatenating short samples to a longer one, resulting in 2.0M data. To further enhance the cross-modality feature alignment, we use the audio in video data when available. We further use the CosyVoice2~\cite{du2024cosyvoice} Text-to-Speech (TTS) method to change a small part of text questions in image and video data into speech following VITA~\cite{fu2024vita}. To make sure the speech is rich in diversity, we select the voice of 5,962 speakers in VoxCeleb2~\cite{chung2018voxceleb2} as the condition for CosyVoice2 to synthesize speech. The training starts with tuning the projector with one quarter of the data and then training the projector and the LLM with LoRA and all data.

In the second streaming finetuning stage, we enhance the model’s streaming question-answering ability and proactive output ability. Thus the data should be annotated with timestamps. In this stage, we use 81k data from MMDuet~\cite{mmduet} with temporal video grounding task, dense captioning task, and multi-answer question answering task, 42k data from ET-Instruct~\cite{liu2024bench} for temporal action localization task and referred video captioning task, and 42k data from EgoTimeQA~\cite{di2024grounded} for general question answering task. We also sample 500k offline data in stage 1 to enrich the dataset. Finally, the training dataset comprises 657k samples. The informative head is trained at this stage.

Finally, we finetune the model on our collected ViSpeak-Instruct dataset, giving the model the ability to mine the instructions in the visual modality and respond to users actively. The resulting model ViSpeak serves as a solid baseline on ViSpeak-Bench.

\begin{table*}[t]
  \centering
  \vspace{-0.6em}
  \resizebox{\linewidth}{!}{
      \begin{tabular}{lcccc|ccccccc|cccccccc|c}
        \hline
        \multirow{2}{*}{Method} & \multirow{2}{*}{Params} & \multirow{2}{*}{Frames} & \multirow{2}{*}{Omni} & \multirow{2}{*}{Streaming} & \multicolumn{7}{c|}{Time Accuracy (\%)} & \multicolumn{8}{c|}{Text Score}  &\multirow{2}{*}{Overall}  \\
        & & & &  & AW & VI & HR & VW & VT & GU  & \textbf{All}& VR & AW & VI & HR & VW & VT & GU  &\textbf{All} & \\
        \hline
        Human (Avg) & - & - & - & - & 70.00 & 100.00 & 90.00 & 92.00 & 96.00 & 98.80 & 91.13 & 4.80 & 2.45 & 4.58 & 3.06 & 5.00 & 5.00 & 2.85 & 3.96 & 3.69 \\
        Human (Max) & - & - & - & - & 70.00 & 100.00 & 100.00 & 100.00 & 100.00 & 100.00 & 95.00 & 5.00 & 2.71 & 5.00 & 3.62 & 5.00 & 5.00 & 3.19 & 4.22 & 4.01 \\
        \hline
        \rowcolor{Gray} \multicolumn{21}{c}{\textbf{Proprietary MLLMs}}\\
        Gemini 1.5 pro~\cite{team2024gemini} & - & - & \checkmark & \ding{55} & 46.00 & 60.00 & 85.00 & 84.00 & 48.00 & 97.00 & 70.00 & 3.03 & 2.34 & 2.93 & 1.36 & 4.66 & 4.68 & 2.07 & 3.01 & 2.19 \\
        GPT-4o~\cite{hurst2024gpt4o} & - & - & \checkmark & \ding{55} & 48.50 & 82.00 & 96.00 & 99.00 & 100.00 & 99.50 & 87.50 & 3.18 & 2.27 & 3.53 & 1.71 & 5.00 & 4.98 & 2.22 & 3.27 & 2.99 \\
        \hline
        \rowcolor{Gray} \multicolumn{21}{c}{\textbf{Open-Source Video MLLMs}}\\
        InternVL-2.5~\cite{chen2024expanding} & 8B & 16 & \ding{55} & \ding{55} & 41.50 & 55.50 & 46.00 & 96.00 & 72.00 & 99.50 & 68.42 & 2.93 & 2.16 & 3.67 & 0.74 & 3.05 & 4.81 & 1.26 & 2.66 & 1.98 \\
        Qwen2.5-VL~\cite{bai2025qwen25vl} & 7B & 1 fps & \ding{55} & \ding{55} & 42.50 & 78.00 & 31.00 & 95.00 & 85.00 & 98.50 & 71.67 & 2.34 & 2.31 & 2.31 & 1.32 & 5.00 & 3.91 & 1.02 & 2.60 & 2.25 \\
        Qwen2.5-VL~\cite{bai2025qwen25vl} & 72B & 1 fps & \ding{55} & \ding{55} & 44.50 & 81.00 & 77.00 & 91.00 & 91.00 & 93.00 & 79.58 & 3.15 & 2.64 & 3.36 & 1.00 & 5.00 & 5.00 & 1.50 & 3.09 & 2.62 \\
        VITA 1.5~\cite{fu2025vita15} & 7B & 1 fps & \checkmark & \ding{55} &  18.00 & 46.00 & 40.00 & 88.00 & 49.00 & 97.50 & 56.42 & 2.40 & 2.08 & 0.57 & 0.85 & 4.57 & 4.49 & 1.18 & 2.31 & 1.54 \\
        Ola~\cite{liu2025ola} & 7B & 1 fps & \checkmark & \ding{55} & 27.00 & 67.00 & 44.00 & 89.00 & 69.00 & 98.50 & 65.75 & 2.95 & 1.81 & 2.67 & 0.55 & 4.71 & 3.67 & 1.52 & 2.55 & 1.86 \\
        FlashVstream~\cite{zhang2024flash} & 7B & 1 fps & \ding{55} & \checkmark & 34.00 & 16.00 & 48.00 & 75.00 & 33.00 & 99.50 & 50.92 & 1.75 & 1.63 & 1.31 & 0.67 & 4.88 & 4.61 & 0.70 & 2.22 & 1.24 \\
        Dispider~\cite{qian2025dispider} & 7B & 16 & \ding{55} & \checkmark  & 38.50	&70.00	&44.00	&69.00&	100.00	&99.50	&70.17& 2.50&	1.75	&4.06	&0.91&	0.61	&2.49&	2.07	&2.06 &1.63\\
        \rowcolor{ours} ViSpeak (Ours, s3) & 7B & 1 fps & \checkmark & \checkmark & 56.50 & 72.00 & 83.00 & 93.00 & 79.00 & 99.00 & 80.42 & 3.75 & 2.63 & 3.84 & 1.07 & 4.95 & 3.15 & 3.36 & 3.25 & 2.76 \\
        \hline
      \end{tabular}
    }
    \vspace{-0.6em}
  \caption{Performance of various MLLMs on ViSpeak-Bench. Results for ViSpeak trained after the third stage are reported. For human evaluation, we invite 5 participants which are not received relevant training to answer 20\% randomly selected questions and we report their average scores and the maximum scores on each subtask.}
  \label{tab:ViSpeakBench_result}
\end{table*}

\section{Experiment}
\vspace{-0.2em}

\subsection{Implementation Detail}

ViSpeak is finetuned from VITA 1.5~\cite{fu2025vita15} due to its high performance on omni-modal data and early open-resourcing, which uses Qwen2.5 7B~\cite{yang2024qwen25} as the LLM and InternViT-300M-448px~\cite{chen2024internvl} as the visual encoder. The audio encoder is designed by VITA itself and has 341M parameters. In the first stage, we first employ a learning
rate of 5e-4 and batch size 256 for MLP adapter pre-training and a learning rate of 1e-4 and batch size 128 for LLM LoRA fintuning. The number of tokens for each image is 256 and the maximum number of images per video is 16. In the second and third stages, the training configurations are the same as those in stage 1 finetuning. However, as streaming video always lasts for a few minutes, we further downsample the image for each frame by a factor of 2, resulting in 64 tokens per image, and increase the maximum number of images per video to 64 accordingly, to extend the context. By default, videos are sampled at 1 fps.  All experiments are conducted on 32 NVIDIA L20 GPUs and the max context length is set to 6,200 due to resource limit. The threshold for the informative head is set to 0.35 for all experiments.

\vspace{-0.2em}
\subsection{Streaming Video Understanding Benchmarks}

In this subsection, we select two large-scale comprehensive streaming video understanding benchmarks for evaluation.

\vspace{0.2em}\noindent{\textbf{Performance on StreamingBench.}} StreamingBench~\cite{lin2024streamingbench} is designed for evaluating real-time visual understanding, omni-source understanding and contextual understanding, which is a comprehensive benchmark for streaming video understanding. It has 18 tasks, 900 videos and 4,500 QA pairs. As shown in \Cref{tab:streamingbench_result}, our ViSpeak model achieves SOTA performance among open-sourced models with only 7B parameters. And the performance is also comparable with GPT-4o, which is a well-known model for its omni-source understanding and interactive ability. And the performance of ViSpeak on omni-source understanding is even higher than GPT-4o (61.60 vs 50.95), demonstrating its outstanding omni-modal comprehensive understanding. Further, with our informative head, our model can speak proactively and get 38.80 scores on PO tasks, while other models should change the proactive problem to an offline one. After the stage 3 finetuning, the proactive output ability is further enhanced and gets 50.80 scores.

\vspace{0.2em}\noindent{\textbf{Performance on OVO-Bench.}} OVO-Bench~\cite{li2025ovo} is designed for evaluating the backward tracing ability, the real-time visual perception ability and the forward active responding ability, which evaluates the model's streaming video understanding capability from a different dimension compared with StreamingBench~\cite{lin2024streamingbench}. It has 12 tasks, 644 videos and 2,800 QA pairs. As shown in \Cref{tab:ovobench_result}, our ViSpeak model also achieves SOTA performance among open-sourced models and the performance is even higher than that of GPT-4o, showing a great ability to handle time-sensitive characteristics in video streaming understanding.

\subsection{ViSpeak-Bench}

On ViSpeak-Bench, we evaluate both representative proprietary and open-source MLLMs. We also conduct a human evaluation as a reference. Results are shown in \Cref{tab:ViSpeakBench_result}.

For human evaluation, we find that in most cases, humans are able to provide appropriate responses at a suitable time and achieve the highest score. But the scores on AW, HR, and GU are relatively low. For Anomaly Warning and Humor Reaction, participants overlook some details in their descriptions, leading to a reduction in scores. And participants sometimes fail to accurately describe the gestures depicted in the videos in the gesture understanding subtask.

During testing MLLMs, we observed existing models perform poorly when not given explicit prompts to indicate the exact expected response type,  because these models are unaware they are in a conversational scenario. To ensure the reliability of the evaluation of these models, we provide clear prompts for different subtasks (see Supplement for more details). With explicit prompts, all models achieve stable performance. Some observations are concluded as follows: 
a) Due to its great interactive ability, GPT-4o performs best among all models.
b) Within open-sourced offline models, Qwen2.5-VL\cite{bai2025qwen25vl} performs best and a larger model can get more reasonable responses.
c) For open-sourced omni-modality models Ola~\cite{liu2025ola} and VITA 1.5~\cite{fu2025vita15}, their performances in both time accuracy and text score are inferior to models like InternVL-2.5~\cite{chen2024expanding} and Qwen2.5-VL~\cite{bai2025qwen25vl}, possibly because they prioritize omni-modality, resulting in a weaker focus on visual understanding.
d) For streaming video LMMs, FlashVstream~\cite{zhang2024flash} and Dispider~\cite{qian2025dispider} still underperform Qwen2.5-VL. We find that FlashVstream tends to speak aggressively, always prior to the actions or events, especially for VI and VT in which the actions are not at the beginning of the video.
Additionally, we also use the same prompts for evaluating MMDuet~\cite{mmduet} and VideoLLM-online~\cite{chen2024videollm}, but we find they can not follow the instructions and simply describe the video, \eg ``You look at the camera.'', which is possibly due to their selected training datasets.

In contrast, without explicit prompts, ViSpeak achieves the highest scores among open-source models, owing to fine-tuning from our strong streaming model. However, we observed that the performance on the anomaly warning and humor reaction subtasks is relatively low, as these tasks exhibit considerable variability in real-world scenarios, and understanding humor is difficult for MLLMs without reasoning ability.

\subsection{Ablation Study}

\begin{table}[t]
  \centering
  \resizebox{\linewidth}{!}{
      \begin{tabular}{lccc}
        \hline
        Method & MME & MVBench & Video-MME \\
        \hline
        VITA 1.5 & 2353.5 (1728.9/624.6) & 53.95 & 58 \\
        \hline
        \rowcolor{ours} Adaptive Sum & 2237.0 (1636.3/600.7) & 54.12 & 55 \\
        Linear & 2283.4 (1685.5/597.9) & 52.95 & 56 \\
        Add & 2292.8 (1691.4/601.4) & 54.27 & 55 \\
        \hline
      \end{tabular}
    }
    \vspace{-0.6em}
  \caption{Ablation studies on input stream combination methods}
  \label{tab:ab1}
\end{table}

\noindent{\textbf{Effect of different combination methods to combine two input streams.}} In the ViSpeak model, we propose to use a two-stream chat template to support interactions within streaming videos. Two input streams (one for the user and one for the agent) are combined into a single one before sending to LLM. We design three types of combination methods: `Adaptive Sum', `Linear' and `Add'. The `Add' method directly adds two streams into a single one along the feature channel dimension. The `Linear' method first concatenates two streams along the feature channel dimension and then uses a linear layer to reduce the dimension. The `Adaptive Sum' method first predicts a weight for each stream and then weighted adds two streams. The intuition behind `Adaptive Sum' is that two inputs may not have equal importance at a specific timestamp. When models are generating responses, they may focus more on their previous output tokens, whereas they pay more attention to user input tokens otherwise. In these experiments, we select MME~\cite{fu2023mme} for image understanding ability evaluation and Video-MME~\cite{fu2024videomme} and MVBench~\cite{li2024mvbench} for the evaluation of video understanding capacity. As shown in \Cref{tab:ab1}, after the first template alignment stage, the model can maintain its offline data understanding ability, achieving a performance comparable to our baseline VITA 1.5. Further, we find that different combination methods also perform similarly and we use the `Adaptive Sum' method by default.

\begin{table}[t]
  \centering
  \resizebox{\linewidth}{!}{
      \begin{tabular}{cccc|cccc|c}
        \hline
        Exp & Head & Joint & Token & Real & Omni & Context & All & PO \\
        \hline
        (a) & LM & \checkmark & $<$seg$>$ & \multirow{3}{*}{73.88} & \multirow{3}{*}{51.70} & \multirow{3}{*}{37.70} & \multirow{3}{*}{60.91} & 30.00 \\
        (b) & inform & \ding{55} & $<$seg$>$ & & & & & 34.80 \\
        (c) & inform & \ding{55} & Visual  & & & & & 36.00 \\
        \hline
        \rowcolor{ours} (d) & inform & \checkmark & Visual & 74.36 & 53.50 & 39.60 & 62.00 & 38.80 \\
        \hline
      \end{tabular}
    }
    \vspace{-0.6em}
  \caption{Ablation studies on the design of visual proactive speaking control. Performances on StreamingBench are reported. `Head' denotes using language modeling head or informative head for prediction. `Joint' denotes whether the head is finetuned with LLM. `Token' means which token is used for prediction.}
  \vspace{-0.6em}
  \label{tab:ab2}
\end{table}

\noindent{\textbf{Effect of different designs to control visual proactive output.}} In this work, we jointly train an informative head with the LLM to control visual proactive output. In \Cref{tab:ab2}, we ablate different designs. In Exp (a) to (c), we first train the model except the informative head. Then, we freeze the LLM and train the informative head in Exp (b) and (c). We find that using the language modeling head for proactive control gets limited performance with only 30.00 scores. Training an informative head following MMDuet~\cite{mmduet} on the frozen LLM can get 34.80 scores. We further find that the last visual token in a segment contains more visual cues than the $<$seg$>$ token so training the informative head based on the visual token can further improve the proactive output score to 36.00. Since the LLM in Exp (a) to (c) are frozen, the performance of other tasks in StreamingBench is the same across these experiments. In Exp (d), we jointly train the informative head with LLM and get the highest proactive output performance. We find that other tasks in StreamingBench are also improved by co-training. We speculate that the informative head makes the model aware of the action boundary thus improving the performance on other tasks.

\begin{table}[t]
  \centering
  \resizebox{\linewidth}{!}{
      \begin{tabular}{lccc}
        \hline
        Dataset & HR (Text Score) & GU (Text Score) & Overall \\
        \hline
        \rowcolor{ours} ViSpeak-Instruct & 1.07 & 3.36 & 2.76 \\
        w/o offline data & 1.02 & 3.17 & 2.70 \\
        \hline
      \end{tabular}
    }
    \vspace{-0.6em}
  \caption{Ablation studies on the offline data in ViSpeak-Instruct. Performance on ViSpeak-Bench are reported.}
  \vspace{-0.6em}
  \label{tab:ab3}
\end{table}

\noindent{\textbf{Effect of the different dataset composition of ViSpeak-Instruct.}} Since there are many kinds of gestures in conversations and the gesture in different contexts has different meanings. To enhance the gesture understanding ability, we use some offline data during training, as well as for humor reactions. As shown in \Cref{tab:ab3}, using offline data can increase the generalization ability and get higher scores.

\begin{table}[t]
  \centering
  \resizebox{\linewidth}{!}{
      \begin{tabular}{l|ccc|c|c}
        \hline
        Model & MME & MVBench & Video-MME & StreamingBench & ViSpeak-Bench \\
        \hline
        ViSpeak (s1) & \cellcolor{ours}2237.0 & \cellcolor{ours}54.12 & \cellcolor{ours}55 & - & - \\
        ViSpeak (s2) & 2051.1 & 49.53 & 58 & \cellcolor{ours}62.00 & - \\
        ViSpeak (s3) & 2181.8 & 53.97 & 60 & \cellcolor{ours}62.58 & \cellcolor{ours}2.76 \\
        \hline
      \end{tabular}
    }
    \vspace{-0.6em}
  \caption{Performance on different benchmarks across different training stages. Results in \colorbox{ours}{purple} are reported in above tables.}
  \vspace{-0.8em}
  \label{tab:ab4}
\end{table}

\noindent{\textbf{Comparisons on the performance across the model in different training stages.}} In this work, we adopt a three-stage finetuning recipe by first finetuning an offline model to a SOTA streaming model and then finetuning for the Visual Instruction Feedback task. In \Cref{tab:ab4}, we find the model can effectively preserve the ability learned from previous stages while progressively learning new skills, demonstrating the superiority of our training recipe.

\section{Conclusion}

In this work, we extend the streaming video understanding problem with a new Visual Instruction Feedback task, which requires the model to respond to visual contents actively. To facilitate research, we define seven key subtasks and collect the ViSpeak-Bench for evaluation and the ViSpeak-Instruct for training. To solve this problem, we first adapt an offline omni-modal LMM to our designed chat template, and then finetuning it to get a SOTA streaming LMM. This model is evaluated on two comprehensive streaming benchmarks and gets GPT-4o-level performance. Finally, we finetune the SOTA streaming model on ViSpeak-Instruct and get the ViSpeak model which serves as a strong baseline on ViSpeak-Bench for future research. We hope our work can provide deeper insights into streaming video understanding and human-agent interaction.

{
    \small
    \bibliographystyle{ieeenat_fullname}
    \bibliography{main}

\begin{thebibliography}{68}
\providecommand{\natexlab}[1]{#1}
\providecommand{\url}[1]{\texttt{#1}}
\expandafter\ifx\csname urlstyle\endcsname\relax
  \providecommand{\doi}[1]{doi: #1}\else
  \providecommand{\doi}{doi: \begingroup \urlstyle{rm}\Url}\fi

\bibitem[Bai et~al.(2025)Bai, Chen, Liu, Wang, Ge, Song, Dang, Wang, Wang, Tang, et~al.]{bai2025qwen25vl}
Shuai Bai, Keqin Chen, Xuejing Liu, Jialin Wang, Wenbin Ge, Sibo Song, Kai Dang, Peng Wang, Shijie Wang, Jun Tang, et~al.
\newblock Qwen2. 5-vl technical report.
\newblock \emph{arXiv preprint arXiv:2502.13923}, 2025.

\bibitem[Chen et~al.(2024{\natexlab{a}})Chen, Lv, Wu, Lin, Song, Gao, Liu, Gao, Mao, and Shou]{chen2024videollm}
Joya Chen, Zhaoyang Lv, Shiwei Wu, Kevin~Qinghong Lin, Chenan Song, Difei Gao, Jia-Wei Liu, Ziteng Gao, Dongxing Mao, and Mike~Zheng Shou.
\newblock Videollm-online: Online video large language model for streaming video.
\newblock In \emph{CVPR}, 2024{\natexlab{a}}.

\bibitem[Chen et~al.(2024{\natexlab{b}})Chen, Li, Dong, Zhang, He, Wang, Zhao, and Lin]{chen2024sharegpt4v}
Lin Chen, Jinsong Li, Xiaoyi Dong, Pan Zhang, Conghui He, Jiaqi Wang, Feng Zhao, and Dahua Lin.
\newblock Sharegpt4v: Improving large multi-modal models with better captions.
\newblock In \emph{ECCV}, 2024{\natexlab{b}}.

\bibitem[Chen et~al.(2024{\natexlab{c}})Chen, Wei, Li, Dong, Zhang, Zang, Chen, Duan, Tang, Yuan, et~al.]{chen2025sharegpt4video}
Lin Chen, Xilin Wei, Jinsong Li, Xiaoyi Dong, Pan Zhang, Yuhang Zang, Zehui Chen, Haodong Duan, Zhenyu Tang, Li Yuan, et~al.
\newblock Sharegpt4video: Improving video understanding and generation with better captions.
\newblock In \emph{NeurIPS}, 2024{\natexlab{c}}.

\bibitem[Chen et~al.(2024{\natexlab{d}})Chen, Wang, Cao, Liu, Gao, Cui, Zhu, Ye, Tian, Liu, et~al.]{chen2024expanding}
Zhe Chen, Weiyun Wang, Yue Cao, Yangzhou Liu, Zhangwei Gao, Erfei Cui, Jinguo Zhu, Shenglong Ye, Hao Tian, Zhaoyang Liu, et~al.
\newblock Expanding performance boundaries of open-source multimodal models with model, data, and test-time scaling.
\newblock \emph{arXiv preprint arXiv:2412.05271}, 2024{\natexlab{d}}.

\bibitem[Chen et~al.(2024{\natexlab{e}})Chen, Wang, Tian, Ye, Gao, Cui, Tong, Hu, Luo, Ma, et~al.]{chen2024far}
Zhe Chen, Weiyun Wang, Hao Tian, Shenglong Ye, Zhangwei Gao, Erfei Cui, Wenwen Tong, Kongzhi Hu, Jiapeng Luo, Zheng Ma, et~al.
\newblock How far are we to gpt-4v? closing the gap to commercial multimodal models with open-source suites.
\newblock \emph{arXiv preprint arXiv:2404.16821}, 2024{\natexlab{e}}.

\bibitem[Chen et~al.(2024{\natexlab{f}})Chen, Wu, Wang, Su, Chen, Xing, Zhong, Zhang, Zhu, Lu, et~al.]{chen2024internvl}
Zhe Chen, Jiannan Wu, Wenhai Wang, Weijie Su, Guo Chen, Sen Xing, Muyan Zhong, Qinglong Zhang, Xizhou Zhu, Lewei Lu, et~al.
\newblock Internvl: Scaling up vision foundation models and aligning for generic visual-linguistic tasks.
\newblock In \emph{CVPR}, 2024{\natexlab{f}}.

\bibitem[Cheng et~al.(2024)Cheng, Leng, Zhang, Xin, Li, Chen, Zhu, Zhang, Luo, Zhao, et~al.]{cheng2024videollama}
Zesen Cheng, Sicong Leng, Hang Zhang, Yifei Xin, Xin Li, Guanzheng Chen, Yongxin Zhu, Wenqi Zhang, Ziyang Luo, Deli Zhao, et~al.
\newblock Videollama 2: Advancing spatial-temporal modeling and audio understanding in video-llms.
\newblock \emph{arXiv preprint arXiv:2406.07476}, 2024.

\bibitem[Chung et~al.(2018)Chung, Nagrani, and Zisserman]{chung2018voxceleb2}
Joon~Son Chung, Arsha Nagrani, and Andrew Zisserman.
\newblock Voxceleb2: Deep speaker recognition.
\newblock \emph{arXiv preprint arXiv:1806.05622}, 2018.

\bibitem[D{\'e}fossez et~al.(2024)D{\'e}fossez, Mazar{\'e}, Orsini, Royer, P{\'e}rez, J{\'e}gou, Grave, and Zeghidour]{defossez2024moshi}
Alexandre D{\'e}fossez, Laurent Mazar{\'e}, Manu Orsini, Am{\'e}lie Royer, Patrick P{\'e}rez, Herv{\'e} J{\'e}gou, Edouard Grave, and Neil Zeghidour.
\newblock Moshi: a speech-text foundation model for real-time dialogue.
\newblock \emph{arXiv preprint arXiv:2410.00037}, 2024.

\bibitem[Di and Xie(2024)]{di2024grounded}
Shangzhe Di and Weidi Xie.
\newblock Grounded question-answering in long egocentric videos.
\newblock In \emph{CVPR}, 2024.

\bibitem[Du et~al.(2024)Du, Wang, Chen, Shi, Lv, Zhao, Gao, Yang, Gao, Wang, et~al.]{du2024cosyvoice}
Zhihao Du, Yuxuan Wang, Qian Chen, Xian Shi, Xiang Lv, Tianyu Zhao, Zhifu Gao, Yexin Yang, Changfeng Gao, Hui Wang, et~al.
\newblock Cosyvoice 2: Scalable streaming speech synthesis with large language models.
\newblock \emph{arXiv preprint arXiv:2412.10117}, 2024.

\bibitem[Epstein et~al.(2020)Epstein, Chen, and Vondrick]{epstein2020oops}
Dave Epstein, Boyuan Chen, and Carl Vondrick.
\newblock Oops! predicting unintentional action in video.
\newblock In \emph{CVPR}, 2020.

\bibitem[Fu et~al.(2023)Fu, Chen, Shen, Qin, Zhang, Lin, Yang, Zheng, Li, Sun, et~al.]{fu2023mme}
Chaoyou Fu, Peixian Chen, Yunhang Shen, Yulei Qin, Mengdan Zhang, Xu Lin, Jinrui Yang, Xiawu Zheng, Ke Li, Xing Sun, et~al.
\newblock Mme: A comprehensive evaluation benchmark for multimodal large language models.
\newblock \emph{arXiv preprint arXiv:2306.13394}, 2023.

\bibitem[Fu et~al.(2024{\natexlab{a}})Fu, Dai, Luo, Li, Ren, Zhang, Wang, Zhou, Shen, Zhang, et~al.]{fu2024videomme}
Chaoyou Fu, Yuhan Dai, Yongdong Luo, Lei Li, Shuhuai Ren, Renrui Zhang, Zihan Wang, Chenyu Zhou, Yunhang Shen, Mengdan Zhang, et~al.
\newblock Video-mme: The first-ever comprehensive evaluation benchmark of multi-modal llms in video analysis.
\newblock \emph{arXiv preprint arXiv:2405.21075}, 2024{\natexlab{a}}.

\bibitem[Fu et~al.(2024{\natexlab{b}})Fu, Lin, Long, Shen, Zhao, Zhang, Dong, Wang, Yin, Ma, et~al.]{fu2024vita}
Chaoyou Fu, Haojia Lin, Zuwei Long, Yunhang Shen, Meng Zhao, Yifan Zhang, Shaoqi Dong, Xiong Wang, Di Yin, Long Ma, et~al.
\newblock Vita: Towards open-source interactive omni multimodal llm.
\newblock \emph{arXiv preprint arXiv:2408.05211}, 2024{\natexlab{b}}.

\bibitem[Fu et~al.(2025{\natexlab{a}})Fu, Lin, Wang, Zhang, Shen, Liu, Li, Long, Gao, Li, et~al.]{fu2025vita15}
Chaoyou Fu, Haojia Lin, Xiong Wang, Yi-Fan Zhang, Yunhang Shen, Xiaoyu Liu, Yangze Li, Zuwei Long, Heting Gao, Ke Li, et~al.
\newblock Vita-1.5: Towards gpt-4o level real-time vision and speech interaction.
\newblock \emph{arXiv preprint arXiv:2501.01957}, 2025{\natexlab{a}}.

\bibitem[Fu et~al.(2025{\natexlab{b}})Fu, Yang, Mo, Yan, Wei, Meng, Xie, and Zheng]{fu2025llmdet}
Shenghao Fu, Qize Yang, Qijie Mo, Junkai Yan, Xihan Wei, Jingke Meng, Xiaohua Xie, and Wei-Shi Zheng.
\newblock Llmdet: Learning strong open-vocabulary object detectors under the supervision of large language models.
\newblock In \emph{CVPR}, 2025{\natexlab{b}}.

\bibitem[Hurst et~al.(2024)Hurst, Lerer, Goucher, Perelman, Ramesh, Clark, Ostrow, Welihinda, Hayes, Radford, et~al.]{hurst2024gpt4o}
Aaron Hurst, Adam Lerer, Adam~P Goucher, Adam Perelman, Aditya Ramesh, Aidan Clark, AJ Ostrow, Akila Welihinda, Alan Hayes, Alec Radford, et~al.
\newblock Gpt-4o system card.
\newblock \emph{arXiv preprint arXiv:2410.21276}, 2024.

\bibitem[Hyun et~al.(2023)Hyun, Sung-Bin, Han, Yu, and Oh]{hyun2023smile}
Lee Hyun, Kim Sung-Bin, Seungju Han, Youngjae Yu, and Tae-Hyun Oh.
\newblock Smile: Multimodal dataset for understanding laughter in video with language models.
\newblock \emph{arXiv preprint arXiv:2312.09818}, 2023.

\bibitem[Jiang et~al.(2024)Jiang, Yang, Xiong, Chen, Zeng, Ren, Zhang, et~al.]{jiang2024chatrex}
Qing Jiang, Yuqin Yang, Yuda Xiong, Yihao Chen, Zhaoyang Zeng, Tianhe Ren, Lei Zhang, et~al.
\newblock Chatrex: Taming multimodal llm for joint perception and understanding.
\newblock \emph{arXiv preprint arXiv:2411.18363}, 2024.

\bibitem[Lai et~al.(2024)Lai, Tian, Chen, Li, Yuan, Liu, and Jia]{lai2024lisa}
Xin Lai, Zhuotao Tian, Yukang Chen, Yanwei Li, Yuhui Yuan, Shu Liu, and Jiaya Jia.
\newblock Lisa: Reasoning segmentation via large language model.
\newblock In \emph{CVPR}, 2024.

\bibitem[Li et~al.(2024{\natexlab{a}})Li, Zhang, Guo, Zhang, Li, Zhang, Zhang, Zhang, Li, Liu, et~al.]{li2024llavaonevision}
Bo Li, Yuanhan Zhang, Dong Guo, Renrui Zhang, Feng Li, Hao Zhang, Kaichen Zhang, Peiyuan Zhang, Yanwei Li, Ziwei Liu, et~al.
\newblock Llava-onevision: Easy visual task transfer.
\newblock \emph{arXiv preprint arXiv:2408.03326}, 2024{\natexlab{a}}.

\bibitem[Li et~al.(2023)Li, Wei, Han, and Fan]{li2023intentqa}
Jiapeng Li, Ping Wei, Wenjuan Han, and Lifeng Fan.
\newblock Intentqa: Context-aware video intent reasoning.
\newblock In \emph{Proceedings of the IEEE/CVF international conference on computer vision}, pages 11963--11974, 2023.

\bibitem[Li et~al.(2024{\natexlab{b}})Li, Wang, He, Li, Wang, Liu, Wang, Xu, Chen, Luo, et~al.]{li2024mvbench}
Kunchang Li, Yali Wang, Yinan He, Yizhuo Li, Yi Wang, Yi Liu, Zun Wang, Jilan Xu, Guo Chen, Ping Luo, et~al.
\newblock Mvbench: A comprehensive multi-modal video understanding benchmark.
\newblock In \emph{CVPR}, 2024{\natexlab{b}}.

\bibitem[Li et~al.(2025{\natexlab{a}})Li, Liu, Zhang, Chen, Li, Li, Liu, Ming, Dong, Pan, et~al.]{li2025baichuan}
Yadong Li, Jun Liu, Tao Zhang, Song Chen, Tianpeng Li, Zehuan Li, Lijun Liu, Lingfeng Ming, Guosheng Dong, Da Pan, et~al.
\newblock Baichuan-omni-1.5 technical report.
\newblock \emph{arXiv preprint arXiv:2501.15368}, 2025{\natexlab{a}}.

\bibitem[Li et~al.(2025{\natexlab{b}})Li, Niu, Miao, Ge, Zhou, He, Dong, Duan, Ding, Qian, et~al.]{li2025ovo}
Yifei Li, Junbo Niu, Ziyang Miao, Chunjiang Ge, Yuanhang Zhou, Qihao He, Xiaoyi Dong, Haodong Duan, Shuangrui Ding, Rui Qian, et~al.
\newblock Ovo-bench: How far is your video-llms from real-world online video understanding?
\newblock In \emph{CVPR}, 2025{\natexlab{b}}.

\bibitem[Li et~al.(2024{\natexlab{c}})Li, Wang, Lin, Tang, Zeng, Hu, and Zheng]{li2024techcoach}
Yuan-Ming Li, An-Lan Wang, Kun-Yu Lin, Yu-Ming Tang, Ling-An Zeng, Jian-Fang Hu, and Wei-Shi Zheng.
\newblock Techcoach: Towards technical keypoint-aware descriptive action coaching.
\newblock \emph{arXiv preprint arXiv:2411.17130}, 2024{\natexlab{c}}.

\bibitem[Lin et~al.(2024)Lin, Fang, Chen, Wan, Luo, Li, Liu, and Sun]{lin2024streamingbench}
Junming Lin, Zheng Fang, Chi Chen, Zihao Wan, Fuwen Luo, Peng Li, Yang Liu, and Maosong Sun.
\newblock Streamingbench: Assessing the gap for mllms to achieve streaming video understanding.
\newblock \emph{arXiv preprint arXiv:2411.03628}, 2024.

\bibitem[Liu et~al.(2023)Liu, Li, Wu, and Lee]{llava}
Haotian Liu, Chunyuan Li, Qingyang Wu, and Yong~Jae Lee.
\newblock Visual instruction tuning.
\newblock In \emph{NeurIPS}, 2023.

\bibitem[Liu et~al.(2024{\natexlab{a}})Liu, Li, Li, and Lee]{liu2024improved}
Haotian Liu, Chunyuan Li, Yuheng Li, and Yong~Jae Lee.
\newblock Improved baselines with visual instruction tuning.
\newblock In \emph{CVPR}, 2024{\natexlab{a}}.

\bibitem[Liu et~al.(2024{\natexlab{b}})Liu, Ma, Qi, Wu, Shan, and Chen]{liu2024bench}
Ye Liu, Zongyang Ma, Zhongang Qi, Yang Wu, Ying Shan, and Chang~Wen Chen.
\newblock Et bench: Towards open-ended event-level video-language understanding.
\newblock In \emph{NeurIPS}, 2024{\natexlab{b}}.

\bibitem[Liu et~al.(2025{\natexlab{a}})Liu, Dong, Liu, Hu, Lu, and Rao]{liu2024oryx}
Zuyan Liu, Yuhao Dong, Ziwei Liu, Winston Hu, Jiwen Lu, and Yongming Rao.
\newblock Oryx mllm: On-demand spatial-temporal understanding at arbitrary resolution.
\newblock In \emph{ICLR}, 2025{\natexlab{a}}.

\bibitem[Liu et~al.(2025{\natexlab{b}})Liu, Dong, Wang, Liu, Hu, Lu, and Rao]{liu2025ola}
Zuyan Liu, Yuhao Dong, Jiahui Wang, Ziwei Liu, Winston Hu, Jiwen Lu, and Yongming Rao.
\newblock Ola: Pushing the frontiers of omni-modal language model with progressive modality alignment.
\newblock \emph{arXiv preprint arXiv:2502.04328}, 2025{\natexlab{b}}.

\bibitem[Materzynska et~al.(2019)Materzynska, Berger, Bax, and Memisevic]{materzynska2019jester}
Joanna Materzynska, Guillaume Berger, Ingo Bax, and Roland Memisevic.
\newblock The jester dataset: A large-scale video dataset of human gestures.
\newblock In \emph{ICCV}, 2019.

\bibitem[Mei et~al.(2024)Mei, Meng, Liu, Kong, Ko, Zhao, Plumbley, Zou, and Wang]{mei2024wavcaps}
Xinhao Mei, Chutong Meng, Haohe Liu, Qiuqiang Kong, Tom Ko, Chengqi Zhao, Mark~D Plumbley, Yuexian Zou, and Wenwu Wang.
\newblock Wavcaps: A chatgpt-assisted weakly-labelled audio captioning dataset for audio-language multimodal research.
\newblock \emph{IEEE/ACM Transactions on Audio, Speech, and Language Processing}, 2024.

\bibitem[Panayotov et~al.(2015)Panayotov, Chen, Povey, and Khudanpur]{panayotov2015librispeech}
Vassil Panayotov, Guoguo Chen, Daniel Povey, and Sanjeev Khudanpur.
\newblock Librispeech: an asr corpus based on public domain audio books.
\newblock In \emph{ICASSP}, 2015.

\bibitem[Peng et~al.(2023)Peng, Li, He, Galley, and Gao]{peng2023instruction}
Baolin Peng, Chunyuan Li, Pengcheng He, Michel Galley, and Jianfeng Gao.
\newblock Instruction tuning with gpt-4.
\newblock \emph{arXiv preprint arXiv:2304.03277}, 2023.

\bibitem[Qian et~al.(2024)Qian, Dong, Zhang, Zang, Ding, Lin, and Wang]{qian2025streaming}
Rui Qian, Xiaoyi Dong, Pan Zhang, Yuhang Zang, Shuangrui Ding, Dahua Lin, and Jiaqi Wang.
\newblock Streaming long video understanding with large language models.
\newblock In \emph{NeurIPS}, 2024.

\bibitem[Qian et~al.(2025)Qian, Ding, Dong, Zhang, Zang, Cao, Lin, and Wang]{qian2025dispider}
Rui Qian, Shuangrui Ding, Xiaoyi Dong, Pan Zhang, Yuhang Zang, Yuhang Cao, Dahua Lin, and Jiaqi Wang.
\newblock Dispider: Enabling video llms with active real-time interaction via disentangled perception, decision, and reaction.
\newblock In \emph{CVPR}, 2025.

\bibitem[Sap et~al.(2019)Sap, Rashkin, Chen, LeBras, and Choi]{sap2019socialiqa}
Maarten Sap, Hannah Rashkin, Derek Chen, Ronan LeBras, and Yejin Choi.
\newblock Socialiqa: Commonsense reasoning about social interactions.
\newblock \emph{arXiv preprint arXiv:1904.09728}, 2019.

\bibitem[Shen et~al.(2024)Shen, Xiong, Zhao, Wu, Chen, Zhu, Liu, Xiao, Varadarajan, Bordes, et~al.]{shen2024longvu}
Xiaoqian Shen, Yunyang Xiong, Changsheng Zhao, Lemeng Wu, Jun Chen, Chenchen Zhu, Zechun Liu, Fanyi Xiao, Balakrishnan Varadarajan, Florian Bordes, et~al.
\newblock Longvu: Spatiotemporal adaptive compression for long video-language understanding.
\newblock \emph{arXiv preprint arXiv:2410.17434}, 2024.

\bibitem[Team et~al.(2024)Team, Georgiev, Lei, Burnell, Bai, Gulati, Tanzer, Vincent, Pan, Wang, et~al.]{team2024gemini}
Gemini Team, Petko Georgiev, Ving~Ian Lei, Ryan Burnell, Libin Bai, Anmol Gulati, Garrett Tanzer, Damien Vincent, Zhufeng Pan, Shibo Wang, et~al.
\newblock Gemini 1.5: Unlocking multimodal understanding across millions of tokens of context.
\newblock \emph{arXiv preprint arXiv:2403.05530}, 2024.

\bibitem[Wang et~al.(2025)Wang, Shan, Shi, Lin, Fei, Tang, Liao, Tang, Huang, and Zheng]{wang2024pargo}
An-Lan Wang, Bin Shan, Wei Shi, Kun-Yu Lin, Xiang Fei, Guozhi Tang, Lei Liao, Jingqun Tang, Can Huang, and Wei-Shi Zheng.
\newblock Pargo: Bridging vision-language with partial and global views.
\newblock In \emph{AAAI}, 2025.

\bibitem[Wang et~al.(2024{\natexlab{a}})Wang, Xu, Cheng, Diao, Zhou, Cao, Wang, Ge, and Huang]{wang2024grounded}
Haibo Wang, Zhiyang Xu, Yu Cheng, Shizhe Diao, Yufan Zhou, Yixin Cao, Qifan Wang, Weifeng Ge, and Lifu Huang.
\newblock Grounded-videollm: Sharpening fine-grained temporal grounding in video large language models.
\newblock \emph{arXiv preprint arXiv:2410.03290}, 2024{\natexlab{a}}.

\bibitem[Wang et~al.(2024{\natexlab{b}})Wang, Yuan, Zhang, and Sun]{wang2024tarsier}
Jiawei Wang, Liping Yuan, Yuchen Zhang, and Haomiao Sun.
\newblock Tarsier: Recipes for training and evaluating large video description models.
\newblock \emph{arXiv preprint arXiv:2407.00634}, 2024{\natexlab{b}}.

\bibitem[Wang et~al.(2024{\natexlab{c}})Wang, Bai, Tan, Wang, Fan, Bai, Chen, Liu, Wang, Ge, et~al.]{wang2024qwen2}
Peng Wang, Shuai Bai, Sinan Tan, Shijie Wang, Zhihao Fan, Jinze Bai, Keqin Chen, Xuejing Liu, Jialin Wang, Wenbin Ge, et~al.
\newblock Qwen2-vl: Enhancing vision-language model's perception of the world at any resolution.
\newblock \emph{arXiv preprint arXiv:2409.12191}, 2024{\natexlab{c}}.

\bibitem[Wang et~al.(2024{\natexlab{d}})Wang, Ren, Luo, Li, Yan, Chen, Wang, Li, Lu, Zhu, et~al.]{asv2}
Weiyun Wang, Yiming Ren, Haowen Luo, Tiantong Li, Chenxiang Yan, Zhe Chen, Wenhai Wang, Qingyun Li, Lewei Lu, Xizhou Zhu, et~al.
\newblock The all-seeing project v2: Towards general relation comprehension of the open world.
\newblock In \emph{ECCV}, 2024{\natexlab{d}}.

\bibitem[Wang et~al.(2024{\natexlab{e}})Wang, Meng, Wang, Liang, Wei, Zhang, and Zhao]{mmduet}
Yueqian Wang, Xiaojun Meng, Yuxuan Wang, Jianxin Liang, Jiansheng Wei, Huishuai Zhang, and Dongyan Zhao.
\newblock Videollm knows when to speak: Enhancing time-sensitive video comprehension with video-text duet interaction format.
\newblock \emph{arXiv preprint arXiv:2411.17991}, 2024{\natexlab{e}}.

\bibitem[Xiao et~al.(2021)Xiao, Shang, Yao, and Chua]{xiao2021next}
Junbin Xiao, Xindi Shang, Angela Yao, and Tat-Seng Chua.
\newblock Next-qa: Next phase of question-answering to explaining temporal actions.
\newblock In \emph{CVPR}, 2021.

\bibitem[Xie et~al.(2024)Xie, Zhang, Zhou, Li, Zhang, Hessel, Yang, and Liu]{xie2024funqa}
Binzhu Xie, Sicheng Zhang, Zitang Zhou, Bo Li, Yuanhan Zhang, Jack Hessel, Jingkang Yang, and Ziwei Liu.
\newblock Funqa: Towards surprising video comprehension.
\newblock In \emph{ECCV}, 2024.

\bibitem[Xie and Wu(2024)]{xie2024miniomni2}
Zhifei Xie and Changqiao Wu.
\newblock Mini-omni2: Towards open-source gpt-4o with vision, speech and duplex capabilities.
\newblock \emph{arXiv preprint arXiv:2410.11190}, 2024.

\bibitem[Xiong et~al.(2025)Xiong, Yang, Yu, Zhuge, Zhang, Zhu, and Lu]{xiong2025streaming}
Haomiao Xiong, Zongxin Yang, Jiazuo Yu, Yunzhi Zhuge, Lu Zhang, Jiawen Zhu, and Huchuan Lu.
\newblock Streaming video understanding and multi-round interaction with memory-enhanced knowledge.
\newblock In \emph{ICLR}, 2025.

\bibitem[Xu et~al.(2024{\natexlab{a}})Xu, Gao, Gan, Chen, Lai, Gang, Kang, and Dehghan]{xu2024slowfast}
Mingze Xu, Mingfei Gao, Zhe Gan, Hong-You Chen, Zhengfeng Lai, Haiming Gang, Kai Kang, and Afshin Dehghan.
\newblock Slowfast-llava: A strong training-free baseline for video large language models.
\newblock \emph{arXiv preprint arXiv:2407.15841}, 2024{\natexlab{a}}.

\bibitem[Xu et~al.(2024{\natexlab{b}})Xu, Jiang, Niu, Deng, Poovendran, Choi, and Lin]{xu2024magpie}
Zhangchen Xu, Fengqing Jiang, Luyao Niu, Yuntian Deng, Radha Poovendran, Yejin Choi, and Bill~Yuchen Lin.
\newblock Magpie: Alignment data synthesis from scratch by prompting aligned llms with nothing.
\newblock \emph{arXiv preprint arXiv:2406.08464}, 2024{\natexlab{b}}.

\bibitem[Yang et~al.(2024)Yang, Yang, Zhang, Hui, Zheng, Yu, Li, Liu, Huang, Wei, et~al.]{yang2024qwen25}
An Yang, Baosong Yang, Beichen Zhang, Binyuan Hui, Bo Zheng, Bowen Yu, Chengyuan Li, Dayiheng Liu, Fei Huang, Haoran Wei, et~al.
\newblock Qwen2. 5 technical report.
\newblock \emph{arXiv preprint arXiv:2412.15115}, 2024.

\bibitem[Yang et~al.(2025{\natexlab{a}})Yang, Bai, Peng, and Wei]{yang2025omni}
Qize Yang, Detao Bai, Yi-Xing Peng, and Xihan Wei.
\newblock Omni-emotion: Extending video mllm with detailed face and audio modeling for multimodal emotion analysis.
\newblock \emph{arXiv preprint arXiv:2501.09502}, 2025{\natexlab{a}}.

\bibitem[Yang et~al.(2025{\natexlab{b}})Yang, Hu, Du, Xue, Qian, Wu, Yang, Dong, and Xu]{yang2025svbench}
Zhenyu Yang, Yuhang Hu, Zemin Du, Dizhan Xue, Shengsheng Qian, Jiahong Wu, Fan Yang, Weiming Dong, and Changsheng Xu.
\newblock Svbench: A benchmark with temporal multi-turn dialogues for streaming video understanding.
\newblock In \emph{ICLR}, 2025{\natexlab{b}}.

\bibitem[Yao et~al.(2024)Yao, Yu, Zhang, Wang, Cui, Zhu, Cai, Li, Zhao, He, et~al.]{yao2024minicpm}
Yuan Yao, Tianyu Yu, Ao Zhang, Chongyi Wang, Junbo Cui, Hongji Zhu, Tianchi Cai, Haoyu Li, Weilin Zhao, Zhihui He, et~al.
\newblock Minicpm-v: A gpt-4v level mllm on your phone.
\newblock \emph{arXiv preprint arXiv:2408.01800}, 2024.

\bibitem[Yu et~al.(2019)Yu, Xu, Yu, Yu, Zhao, Zhuang, and Tao]{yu2019activitynetqa}
Zhou Yu, Dejing Xu, Jun Yu, Ting Yu, Zhou Zhao, Yueting Zhuang, and Dacheng Tao.
\newblock Activitynet-qa: A dataset for understanding complex web videos via question answering.
\newblock In \emph{AAAI}, 2019.

\bibitem[Zadeh et~al.(2019)Zadeh, Chan, Liang, Tong, and Morency]{zadeh2019social}
Amir Zadeh, Michael Chan, Paul~Pu Liang, Edmund Tong, and Louis-Philippe Morency.
\newblock Social-iq: A question answering benchmark for artificial social intelligence.
\newblock In \emph{CVPR}, 2019.

\bibitem[Zhang et~al.(2024{\natexlab{a}})Zhang, Wang, Tang, Liu, Feng, Dai, and Jin]{zhang2024flash}
Haoji Zhang, Yiqin Wang, Yansong Tang, Yong Liu, Jiashi Feng, Jifeng Dai, and Xiaojie Jin.
\newblock Flash-vstream: Memory-based real-time understanding for long video streams.
\newblock \emph{arXiv preprint arXiv:2406.08085}, 2024{\natexlab{a}}.

\bibitem[Zhang et~al.(2024{\natexlab{b}})Zhang, Xu, Wang, Zuo, Huang, Gao, Zhang, Yu, and Sang]{zhang2024holmes}
Huaxin Zhang, Xiaohao Xu, Xiang Wang, Jialong Zuo, Xiaonan Huang, Changxin Gao, Shanjun Zhang, Li Yu, and Nong Sang.
\newblock Holmes-vau: Towards long-term video anomaly understanding at any granularity.
\newblock \emph{arXiv preprint arXiv:2412.06171}, 2024{\natexlab{b}}.

\bibitem[Zhang et~al.(2024{\natexlab{c}})Zhang, Dong, Cao, Zang, Qian, Wei, Chen, Li, Niu, Ding, et~al.]{zhang2024omnilive}
Pan Zhang, Xiaoyi Dong, Yuhang Cao, Yuhang Zang, Rui Qian, Xilin Wei, Lin Chen, Yifei Li, Junbo Niu, Shuangrui Ding, et~al.
\newblock Internlm-xcomposer2. 5-omnilive: A comprehensive multimodal system for long-term streaming video and audio interactions.
\newblock \emph{arXiv preprint arXiv:2412.09596}, 2024{\natexlab{c}}.

\bibitem[Zhang et~al.(2024{\natexlab{d}})Zhang, Dong, Zang, Cao, Qian, Chen, Guo, Duan, Wang, Ouyang, et~al.]{zhang2024internlm}
Pan Zhang, Xiaoyi Dong, Yuhang Zang, Yuhang Cao, Rui Qian, Lin Chen, Qipeng Guo, Haodong Duan, Bin Wang, Linke Ouyang, et~al.
\newblock Internlm-xcomposer-2.5: A versatile large vision language model supporting long-contextual input and output.
\newblock \emph{arXiv preprint arXiv:2407.03320}, 2024{\natexlab{d}}.

\bibitem[Zhang et~al.(2024{\natexlab{e}})Zhang, Li, Liu, Lee, Gui, Fu, Feng, Liu, and Li]{zhang2024llavanextvideo}
Yuanhan Zhang, Bo Li, haotian Liu, Yong~jae Lee, Liangke Gui, Di Fu, Jiashi Feng, Ziwei Liu, and Chunyuan Li.
\newblock Llava-next: A strong zero-shot video understanding model, 2024{\natexlab{e}}.

\bibitem[Zhang et~al.(2024{\natexlab{f}})Zhang, Wu, Li, Li, Ma, Liu, and Li]{zhang2024video}
Yuanhan Zhang, Jinming Wu, Wei Li, Bo Li, Zejun Ma, Ziwei Liu, and Chunyuan Li.
\newblock Video instruction tuning with synthetic data.
\newblock \emph{arXiv preprint arXiv:2410.02713}, 2024{\natexlab{f}}.

\bibitem[Zhao et~al.(2025)Zhao, Yang, Peng, Bai, Yao, Sun, Chen, Fu, Wei, Bo, et~al.]{zhao2025humanomni}
Jiaxing Zhao, Qize Yang, Yixing Peng, Detao Bai, Shimin Yao, Boyuan Sun, Xiang Chen, Shenghao Fu, Xihan Wei, Liefeng Bo, et~al.
\newblock Humanomni: A large vision-speech language model for human-centric video understanding.
\newblock \emph{arXiv preprint arXiv:2501.15111}, 2025.

\end{thebibliography}
}

\newpage

\section{More Details for ViSpeak-Bench and ViSpeak-Instruct}

\begin{table}[t]
  \centering
      \begin{tabular}{l|l}
        \hline
        Datasets & License \\
        \hline
        OOPS~\cite{epstein2020oops} & CC BY-NC-SA 4.0 \\
        FunQA~\cite{xie2024funqa} & CC BY-NC-SA 4.0 \\
        SocialIQA~\cite{sap2019socialiqa} & MIT \\
        HIVAU~\cite{zhang2024holmes} & MIT \\
        Social-IQ~\cite{zadeh2019social} & MIT \\
        IntentQA~\cite{li2023intentqa} & N/A \\
        Jester~\cite{materzynska2019jester} & N/A \\
        SMILE~\cite{hyun2023smile} & N/A \\
        \hline
      \end{tabular}
    \vspace{-0.6em}
  \caption{License of source datasets in ViSpeak-Bench and ViSpeak-Instruct.}
  \vspace{-0.8em}
  \label{tab:license}
\end{table}

\begin{figure}[t]
  \centering
  \includegraphics[width=1\linewidth]{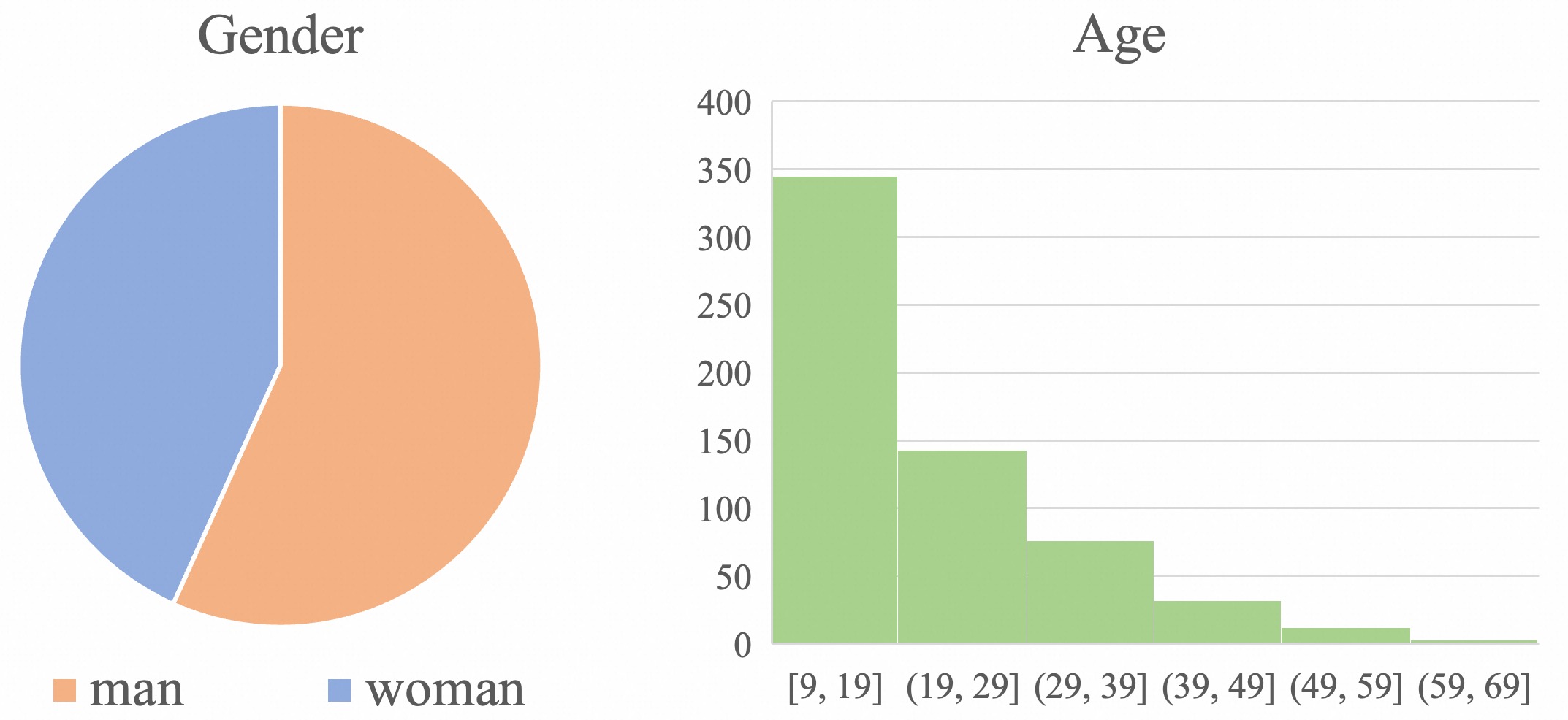}
  \caption{Statistics on participants who recorded videos. The participants comprised nearly equal numbers of males and females, with ages ranging from 10 to 70 years.}
  \vspace{-0.6em}
  \label{fig:personnel_information}
\end{figure}

\subsection{Licenses}
The self-collected videos in our ViSpeak-Bench and ViSpeak-Instruct are provided to the community under \textbf{CC BY-NC-SA 4.0} license. By downloading our dataset from our website or other sources, the user agrees to adhere to the terms of \textbf{CC BY-NC-SA 4.0} and licenses of other source datasets. Licenses of the source datasets are listed in \Cref{tab:license}.

\subsection{Participants in Collecting Videos}
To collect the ViSpeak-Bench and ViSpeak-Instruct datasets, we recruit a team of 610 people (346 men and 264 women) with an age ranging from 10 to 70 years old from 5 provinces, summarized in \Cref{fig:personnel_information}. We obtained signed agreements from each participant to ensure that their data can be utilized by the community.

\subsection{Prompts for Dataset Construction}
During the data collection procedure, we use GPT-4o~\cite{hurst2024gpt4o} to reformulate the responses and generate conversation scripts. Since original datasets have high-quality annotations, we directly use these annotations as conditions, which greatly decreases the difficulty for GPT-4o to translate. The reformulated responses contain two parts: the first one is what action or event happens and the second part is some reasonable responses toward the action or event. Prompts for Anomaly Warning and Humor Reaction are displayed as follows.

\begin{tcolorbox}[colback=gray!10,colframe=black!80,title=The prompt for reformulating the responses in HIVAU dataset]
Suppose you are a helpful AI chatbot that will give the user some advice based on any anomalous situations. You should first identify whether an anomaly event exists. If it does, give the user some advice in a sentence and in a conversational tone assuming the event has actually happened. The output should be in a dict format, like \{`anomaly': 0, `advice': None\} or \{`anomaly': 1, `advice': `Your advice'\}, where 0 indicates no anomaly event and 1 indicates an anomaly event.

Description: \{caption\} 

Output:
\end{tcolorbox}

\begin{tcolorbox}[colback=gray!10,colframe=black!80,title=The prompt for reformulating the responses in OOPS dataset]
Suppose you are a helpful AI chatbot that will give the user some advice based on the given unintentional situations. Assume you have seen the situation and remind the user. You should first describe the situation and give the user some advice in a sentence and in a conversational tone.

Description: \{caption\} 

Output:
\end{tcolorbox}

\begin{tcolorbox}[colback=gray!10,colframe=black!80,title=The prompt for reformulating the responses in FunQA dataset]
Change the input to a conversational tone as if you are talking to someone about the scene you are watching now. Do not output imaginary contents. 

Description: \{caption\} 

Output:
\end{tcolorbox}

For the Gesture Understanding task, we manually select 10 common gestures from the Jester~\cite{materzynska2019jester} dataset and the other 10 gestures collected by ourselves. Gestures from Jester are ``Swiping Right'', ``Swiping Down'', ``Swiping Left'', ``Swiping Up'', ``Pulling Hand In'', ``Pushing Hand Away'', ``Zooming Out With Full Hand'', ``Zooming In With Full Hand'', ``Thumb Down'', and ``Thumb Up''. Our self-collected gestures are ``Zero'', ``One'', ``Two'', ``Three'', ``Four'', ``Five'', ``Victory'', ``Finger Heart 1'', ``Finger Heart 2'', and ``OK''. Note that many gestures are similar, for example ``Two'' is similar to ``Victory'', ``Three'' is similar to ``OK''. The meaning of gestures varies in different contexts. Thus, we use GPT-4o to generate a wide variety of scripts for video recording. The prompt is shown below.

\begin{tcolorbox}[colback=gray!10,colframe=black!80,title=The prompt for generating gesture understanding scripts]
Suppose you are talking to a user. Your task is to generate a reasonable conversation context for a gesture from the user. For example, suppose the gesture is `number 5', a reasonable context is \{`human': `Can you share something with me?', `gpt': `I was just looking at how many hours you usually spend on your hobbies each week. How many do you think it is?', `human': `number 5', `gpt': `Your gesture is the number 5. That's great! It sounds like you really dedicate some solid time to your hobbies. What do you enjoy doing the most during those hours?'\}. In the first round, the user start the conversation. Then, in the second round, you should start with various topics. The input gesture in the third round is the feedback from the user. After receiving the feedback from the user, you should first point out the gesture and then generate friendly or helpful feedback in the last round. Note that conversations should be unrelated to a specific environment, but it should be highly reasonable to perform the gesture in this context. You should answer the question following the template in the example.

Gesture: \{gesture\} 

Output:
\end{tcolorbox}

For scripts in the visual termination and visual interruption tasks, we select QA pairs from GPT-4-LLM~\cite{peng2023instruction}, with long ones for Visual Interruption and short ones for Visual Termination.

\subsection{Evaluation Method for Offline Models on ViSpeak-Bench} 
\label{offline_evaluation}
Since most existing Large Video Language Models are offline models, we change the proactive output problems in our ViSpeak-Bench into offline ones following StreamingBench~\cite{lin2024streamingbench} and OVO-Bench~\cite{li2025ovo}. The evaluation is broken into a two-step evaluation. In the first step, we will inquire the model whether it is an appropriate time to provide a response iteratively at each timestamp to find an appropriate time for response. The sub-video from the beginning till now is used as if it is a full video. In the second step, the model generates the actual responses based on the context up to now. Further, since existing offline models are not finetuned on our ViSpeak-Instruct, they can not generate proper responses without explicit prompts and the prompts we used are as follows:

\noindent For \textbf{Gesture Understanding}, the prompts are:
\begin{itemize}
    \item \textit{Step 1}: You're watching a video. At this moment in the video, is there any gesture being made in the video? You can only answer yes or no.
    \item \textit{Step 2}: What gesture did the person in the video make, and what does it signify when considering the context of the preceding conversation?
\end{itemize}

\noindent For \textbf{Visual Wake-Up}, the prompts are:
\begin{itemize}
    \item \textit{Step 1}: You're watching a video. At this moment in the video, is there any gesture/action being made in the video? You can only answer yes or no.
    \item \textit{Step 2}: When you see greeting gesture, what should you respond to me? Directly output your response.
\end{itemize}

\noindent For \textbf{Visual Termination}, the prompts are:
\begin{itemize}
    \item \textit{Step 1}: You're watching a video. At this moment in the video, is there any gesture/action being made in the video? You can only answer yes or no.
    \item \textit{Step 2}: When you see the goodbye gesture, what should you respond to me? Directly output your response.
\end{itemize}

\noindent For \textbf{Visual Interruption}, the prompts are:
\begin{itemize}
    \item \textit{Step 1}: You're watching a video. At this moment in the video, is there any gesture/action being made in the video? You can only answer yes or no.
    \item \textit{Step 2}: When you see the body language or gesture that indicates interruption, you should say stop. What should you respond to me now? Directly output your response.
\end{itemize}

\noindent For \textbf{Anomaly Warning}, the prompts are:
\begin{itemize}
    \item \textit{Step 1}: You're watching a video. At this moment in the video, is there anything unusual happening in the video? You can only answer yes or no.
    \item \textit{Step 2}: What unusual events occur in this video, and what is your suggestion based on these observations?
\end{itemize}

\noindent For \textbf{Humor Reaction}, the prompts are:
\begin{itemize}
    \item \textit{Step 1}: You're watching a video. At this moment in the video, is there anything funny happening in the video? You can only answer yes or no.
    \item \textit{Step 2}: What interesting events occurred in the video, and why?
\end{itemize}

\subsection{Evaluation Prompts for ViSpeak-Bench} 

As most of questions in our ViSpeak-Bench are open-ended questions. Thus, we utilize GPT-4o as the judge for evaluation. The prompts for each subtask are shown as follows.

\begin{tcolorbox}[colback=gray!10,colframe=black!80,title=The prompt for Visual Wake-Up]
Determine if the GPT text expresses greeting intent. 

Use a 0-5 scale where 0 indicates no greeting signals and 5 shows explicit welcoming language.

5 = Explicit welcoming phrases (``Hello!",``How can I assist?")

4 = Contextual opening (Matching time-based greetings like ``Good morning")

3 = Service-oriented initiation (``What can I help with?")

2 = Neutral conversational starters (``Shall we begin?")

1 = Implicit opening cues (Single welcoming emoji)

0 = No greeting/dialogue initiation detected
\\

DO NOT PROVIDE ANY OTHER OUTPUT TEXT OR EXPLANATION.

DO NOT INCLUDE ANY MARKDOWN FORMAT.

Only provide the score and reason in JSON format. For example, \{``score": 3, ``reason": ``this is a reason for the score"\}
\end{tcolorbox}

\begin{tcolorbox}[colback=gray!10,colframe=black!80,title=The prompt for Visual Termination]
Determine if the GPT text contains intent to conclude the conversation. 

Use a 0-5 scale where 0 indicates no termination signals and 5 shows explicit closure intent.

5 = Contains explicit closure phrases (e.g., ``Goodbye", ``End of discussion")

4 = Implies conclusion with summary + sign-off (e.g., ``In summary...feel free to ask more")

3 = Neutral statement with closing tendency (e.g., ``Let me know if needed")

2 = Ambiguous transitional phrasing (e.g., ``Anyway...")

1 = Contains weak closure cues (e.g., single emoji/period)

0 = No termination indicators, continues dialog flow
\\

DO NOT PROVIDE ANY OTHER OUTPUT TEXT OR EXPLANATION.

DO NOT INCLUDE ANY MARKDOWN FORMAT.

Only provide the score and reason in JSON format. For example, \{``score": 3, ``reason": ``this is a reason for the score"\}
\end{tcolorbox}

\begin{tcolorbox}[colback=gray!10,colframe=black!80,title=The prompt for Visual Interruption]
Determine if the GPT text contains cessation intent (stopping speaking/concluding). 

Use a 0-5 scale where 0 indicates continued engagement and 5 shows explicit termination.

5 = Explicit acknowledgment (``I'll stop here") + closure

4 = Implicit cessation recognition (``Unless you need more...") 

3 = Passive-agreement cessation (``As requested, ending now")

2 = Ambiguous continuation (``...but") after stop cue

1 = Partial acknowledgment without closure

0 = Complete continuation disregard
\\

DO NOT PROVIDE ANY OTHER OUTPUT TEXT OR EXPLANATION.

DO NOT INCLUDE ANY MARKDOWN FORMAT.

Only provide the score and reason in JSON format. For example, \{``score": 3, ``reason": ``this is a reason for the score"\}

\end{tcolorbox}

\begin{tcolorbox}[colback=gray!10,colframe=black!80,title=The prompt for Anomaly Warning]
Evaluate video anomaly response from GPT with the following metric, taking into account the total score of 5 points, 
with separate scores for Description Consistency between ground truth and Advice Rationality:
\\

Description Consistency between Ground Truth:

3 = Core elements match without errors

2 = Core elements match with minor errors

1 = Superficial match

0 = Key element errors or contradictory
\\

Advice Rationality:

2 = Actionable \& Safe \& Logically sound

1 = Generally appropriate

0 = Dangerous/hallucinated
\\

DO NOT PROVIDE ANY OTHER OUTPUT TEXT OR EXPLANATION.

DO NOT INCLUDE ANY MARKDOWN FORMAT.

Only provide the Total score and reason in JSON format. For example, \{``description": 3, ``advice": 2, ``reason": ``this is a reason for the score"\}

\end{tcolorbox}

\begin{tcolorbox}[colback=gray!10,colframe=black!80,title=The prompt for Humor Reaction]
Evaluate alignment between Ground Truth and GPT Text regarding humorous event descriptions. 

5 = Perfect match in humor and delivery

4 = Preserves main humor, but with minor changes to the story or details

3 = Only partial humor retention with some deviations

2 = Only partial humor retention and some important parts are missing

1 = Superficial similarity only

0 = No comedic correlation
\\

DO NOT PROVIDE ANY OTHER OUTPUT TEXT OR EXPLANATION.

DO NOT INCLUDE ANY MARKDOWN FORMAT.

Only provide the score and reason in JSON format. For example, \{``score": 3, ``reason": ``this is a reason for the score"\}

\end{tcolorbox}

\begin{tcolorbox}[colback=gray!10,colframe=black!80,title=The prompt for Gesture Understanding]
Evaluate gesture response from GPT with the following metric, taking into account the total score of 5 points, 
with separate scores for gesture recognition and contextual appropriateness of the response:

Gesture recognition:

3 = Precise gesture identification

2 = Ambiguous gesture reference

1 = No explicit mention of gestures

0 = Hallucinated/non-existent gesture
\\

Contextual appropriateness:

2 = Natural integration with dialogue

1 = Generic but relevant response

0 = Irrelevant/contradictor response
\\

[Dialogue History] provided for context

[Gesture] is the ground truth

[Contextual Reference Text] as a reference, but does not have to match exactly
\\

DO NOT PROVIDE ANY OTHER OUTPUT TEXT OR EXPLANATION.

DO NOT INCLUDE ANY MARKDOWN FORMAT.

Only provide the Total score and reason in JSON format. For example, \{``description": 3, ``advice": 2, ``reason": ``this is a reason for the score"\}

\end{tcolorbox}

\subsection{Examples of Each Subtask}

\begin{figure*}[t]
  \centering
  \includegraphics[width=0.8\linewidth]{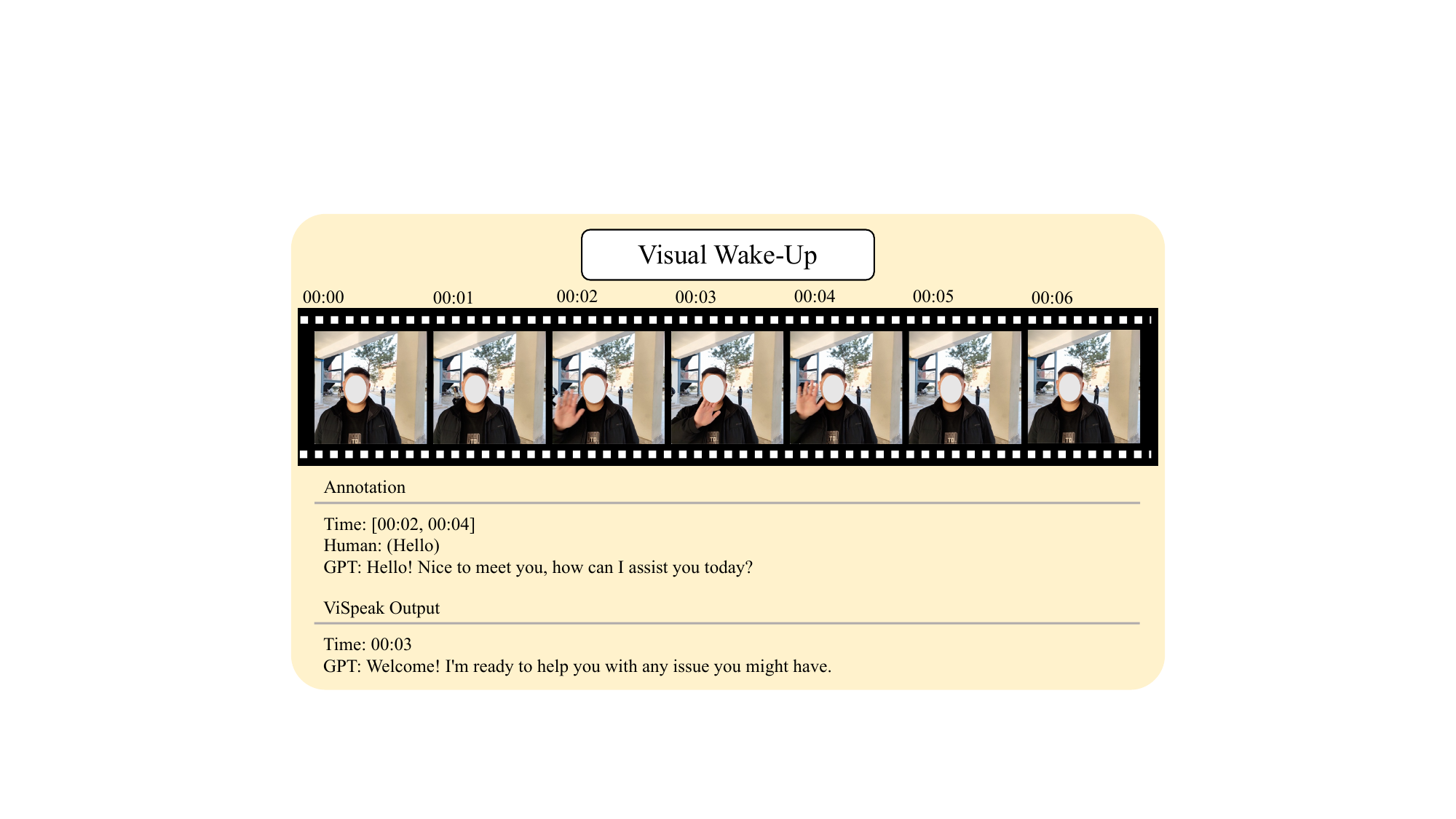}
  \caption{Examples of Visual Wake-Up in ViSpeak-Bench and the corresponding output by ViSpeak.}
  \vspace{-0.6em}
  \label{fig:wakeup}
\end{figure*}

\begin{figure*}[t]
  \centering
  \includegraphics[width=0.8\linewidth]{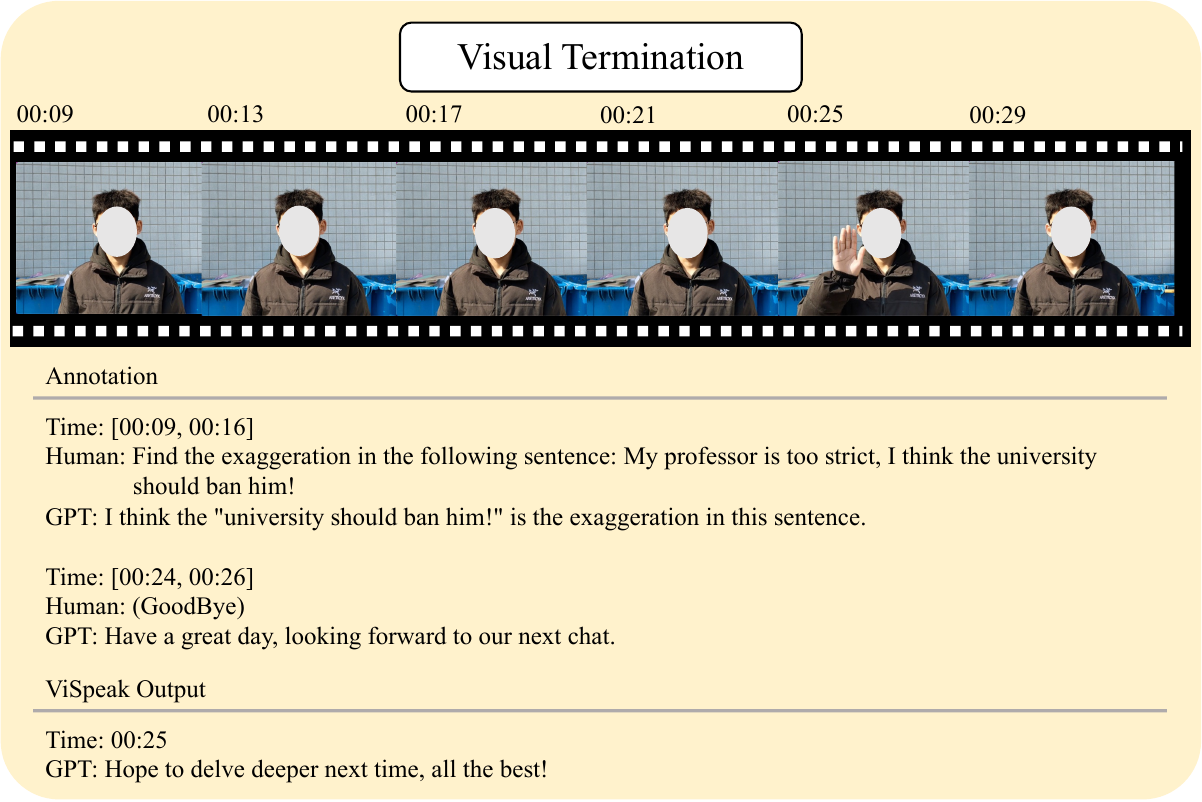}
  \caption{Examples of Visual Termination in ViSpeak-Bench and the corresponding output by ViSpeak. The first round conversation is used as context.}
  \vspace{-0.6em}
  \label{fig:goodbye}
\end{figure*}

\begin{figure*}[t]
  \centering
  \includegraphics[width=0.8\linewidth]{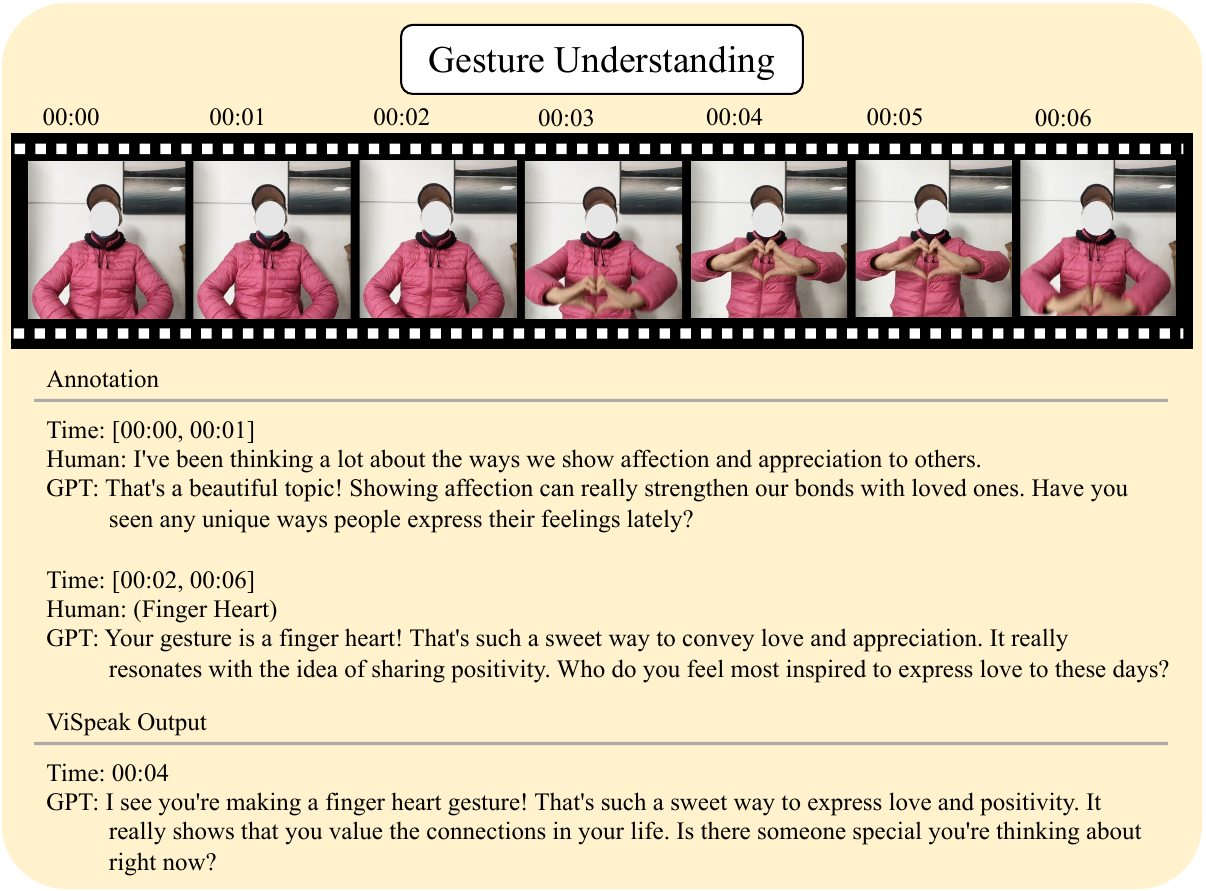}
  \caption{Examples of Gesture Understanding in ViSpeak-Bench and the corresponding output by ViSpeak. The first round conversation is used as context.}
  \vspace{-0.6em}
  \label{fig:gesture}
\end{figure*}

\begin{figure*}[t]
  \centering
  \includegraphics[width=0.8\linewidth]{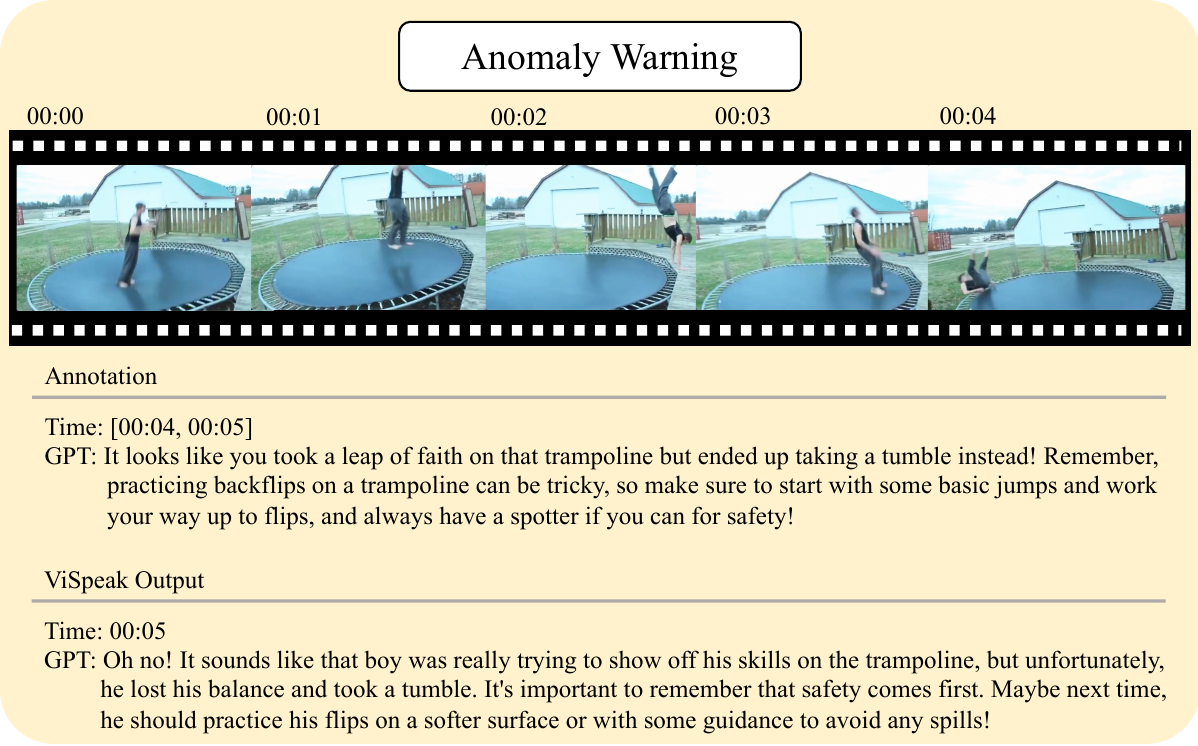}
  \caption{Examples of Anomaly Warning in ViSpeak-Bench and the corresponding output by ViSpeak.}
  \vspace{-0.6em}
  \label{fig:anomaly}
\end{figure*}

\begin{figure*}[t]
  \centering
  \includegraphics[width=0.8\linewidth]{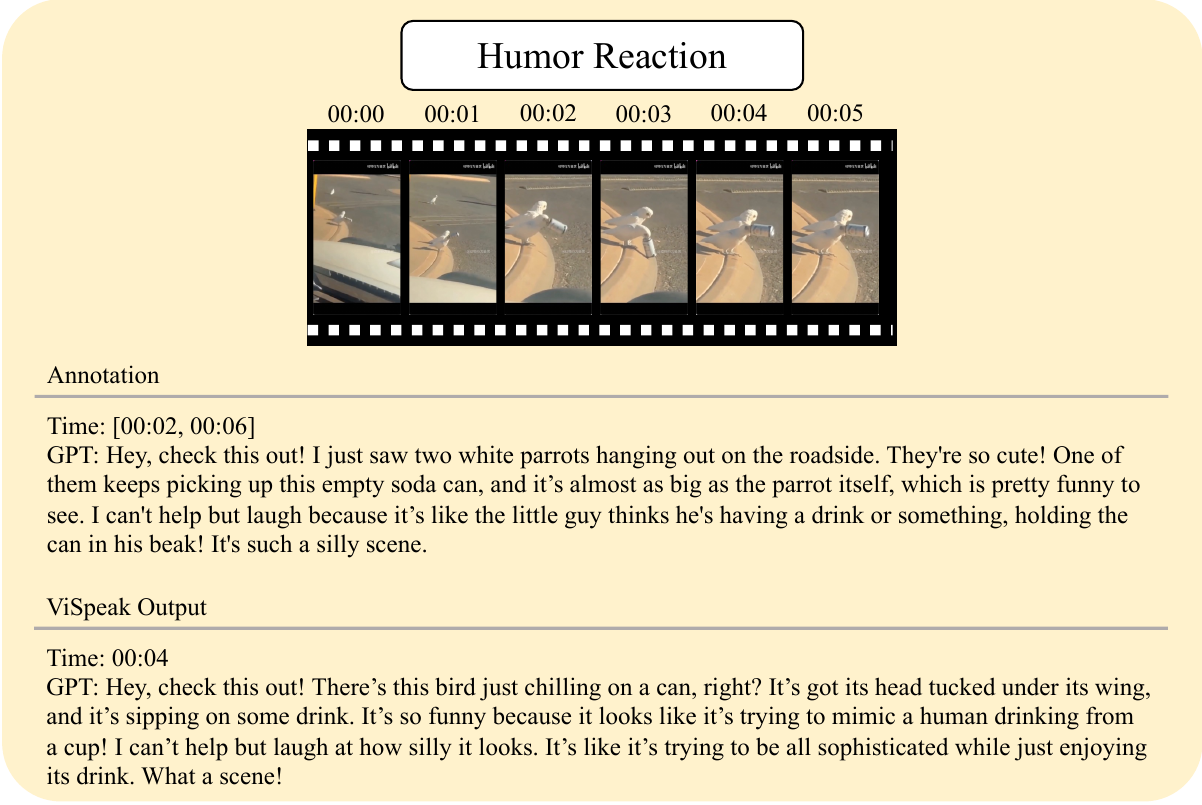}
  \caption{Examples of Humor Reaction in ViSpeak-Bench and the corresponding output by ViSpeak.}
  \vspace{-0.6em}
  \label{fig:humor}
\end{figure*}

\begin{figure*}[t]
  \centering
  \includegraphics[width=0.8\linewidth]{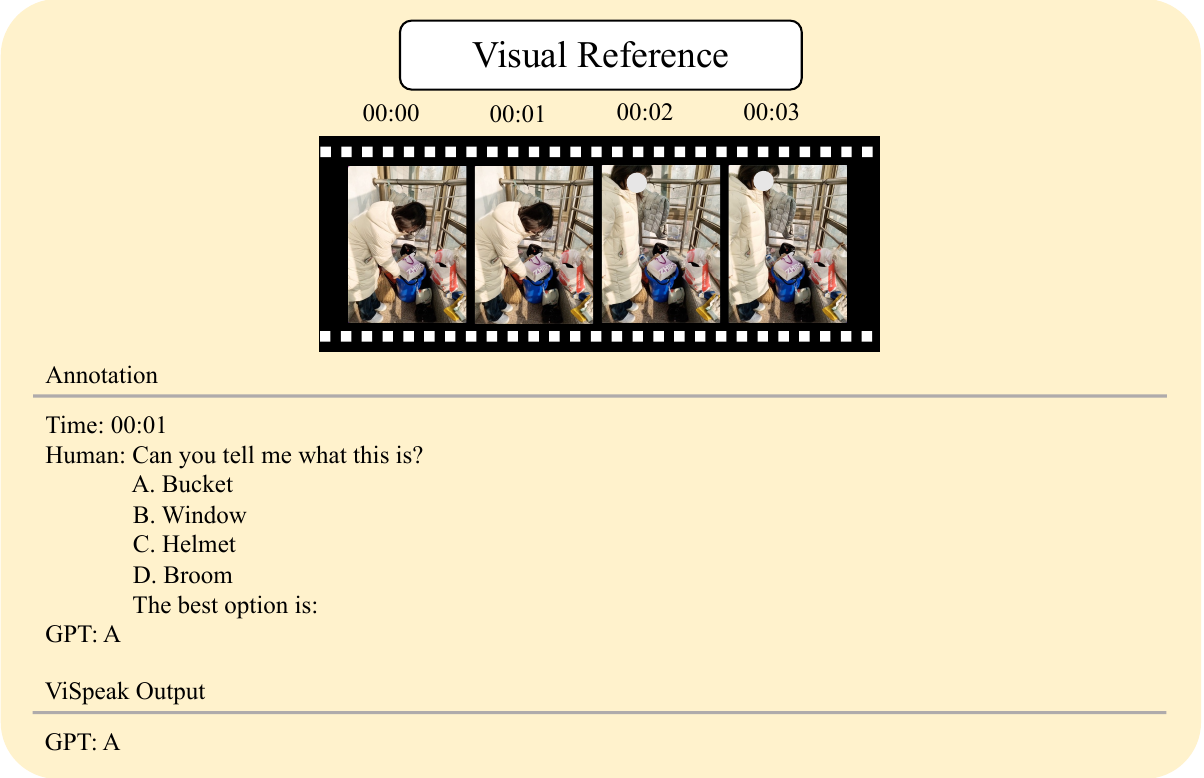}
  \caption{Examples of Visual Reference in ViSpeak-Bench and the corresponding output by ViSpeak.}
  \vspace{-0.6em}
  \label{fig:reference}
\end{figure*}

From \Cref{fig:wakeup} to \Cref{fig:reference}, we visualize some samples in each task, each of which is annotated with accurate timestamps and a referenced response. We also visualize the outputs from our ViSpeak model.

\subsection{Examples of Self-Annotated Gesture Understanding Data}

\begin{figure*}[t]
  \centering
  \includegraphics[width=0.8\linewidth]{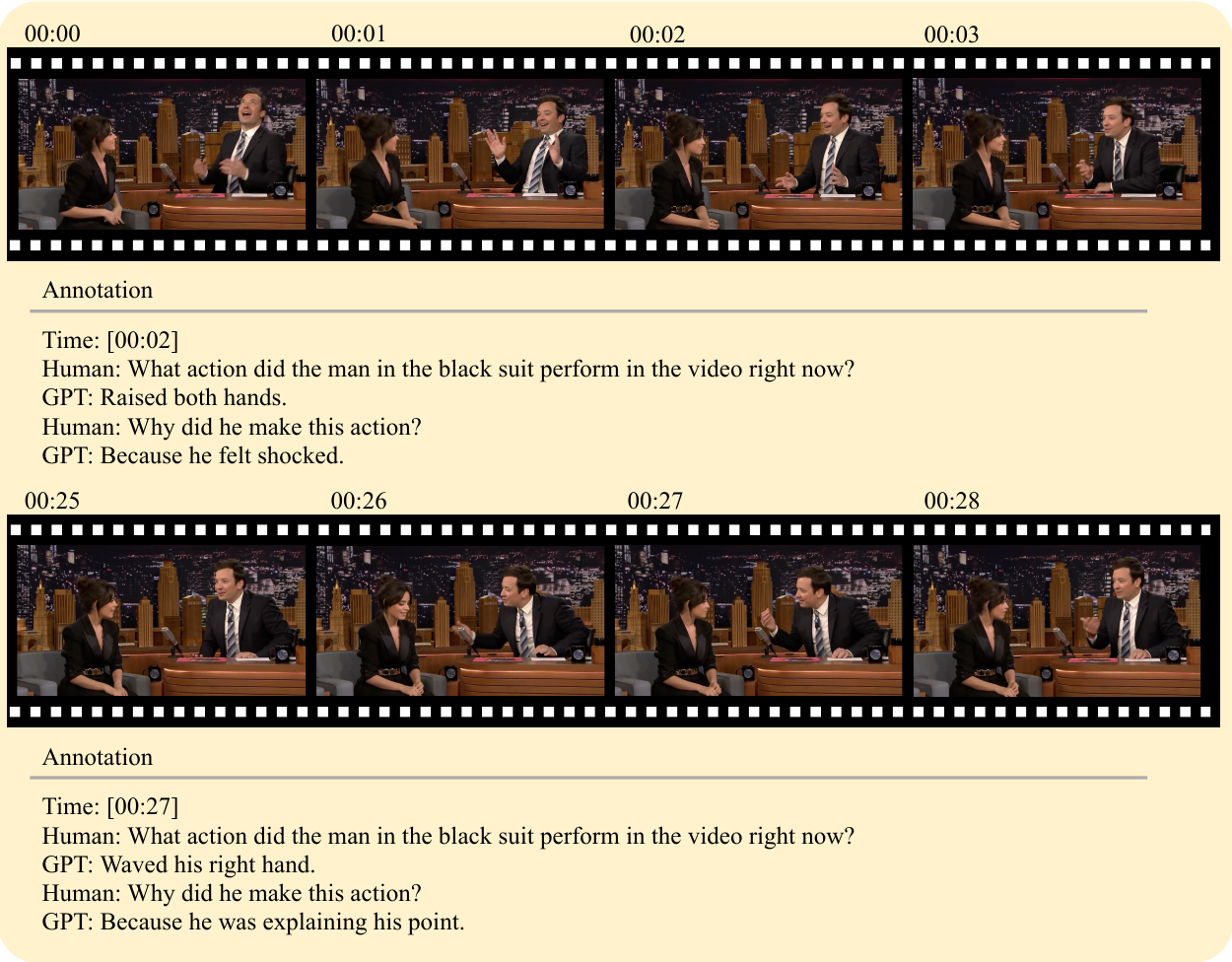}
  \caption{Examples of our self-annotated data for gesture understanding.}
  \vspace{-0.6em}
  \label{fig:socialiq}
\end{figure*}

In \Cref{fig:socialiq}, we visualize some examples of self-annotated gesture understanding data. Each sample is annotated with two questions: the first one is to ask what the gesture is and the second one is to ask the meaning of the gesture. Gestures in natural conversations greatly enhance the diversity of our dataset.

\subsection{Evaluation of ViSpeak on Visual Interruption}

\begin{figure*}[t]
  \centering
  \includegraphics[width=0.8\linewidth]{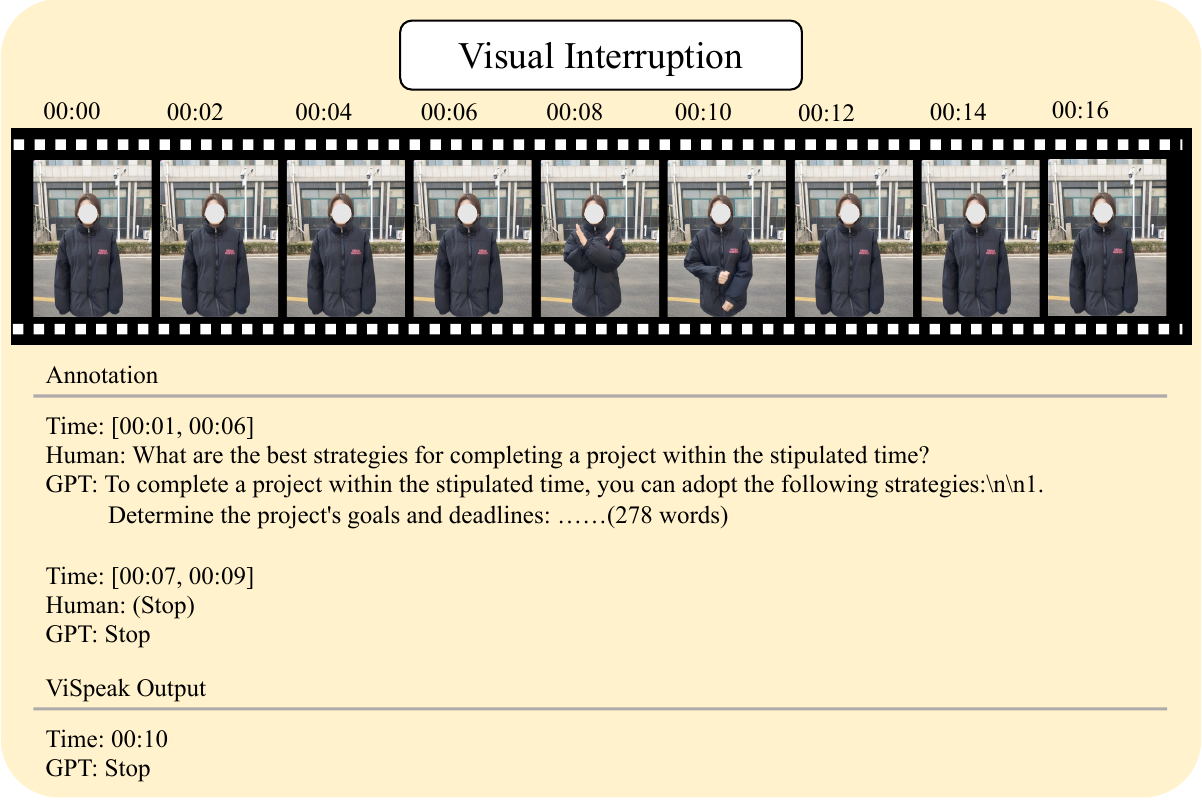}
  \caption{Examples of Visual Interruption in ViSpeak-Bench and the corresponding output by ViSpeak. The first round conversation is used as context.}
  \vspace{-0.6em}
  \label{fig:interrpt}
\end{figure*}

Since recent LMMs can not be interrupted by visual instructions, we actually do not evaluate their ability to be interrupted. As illustrated in \Cref{offline_evaluation}, we simplify the problem to recognize the stop gesture. But when evaluating our ViSpeak, we use the following methods to evaluate the ability to be interrupted. Taking \Cref{fig:interrpt} as an example, we assume that the question from the user arises at 00:06. Then, we directly use the long reply from the annotation files as responses to prevent the model-generated replies from being too short to be interrupted. We replace the predicted token in the next token prediction with the token in the long reply until a ``$\Downarrow$'' token is predicted on a $<$seg$>$ token, which means the model is interrupted.

\subsection{Failure Case and Analysis}

\begin{figure*}[t]
  \centering
  \vspace{-4em}
  \includegraphics[width=0.8\linewidth]{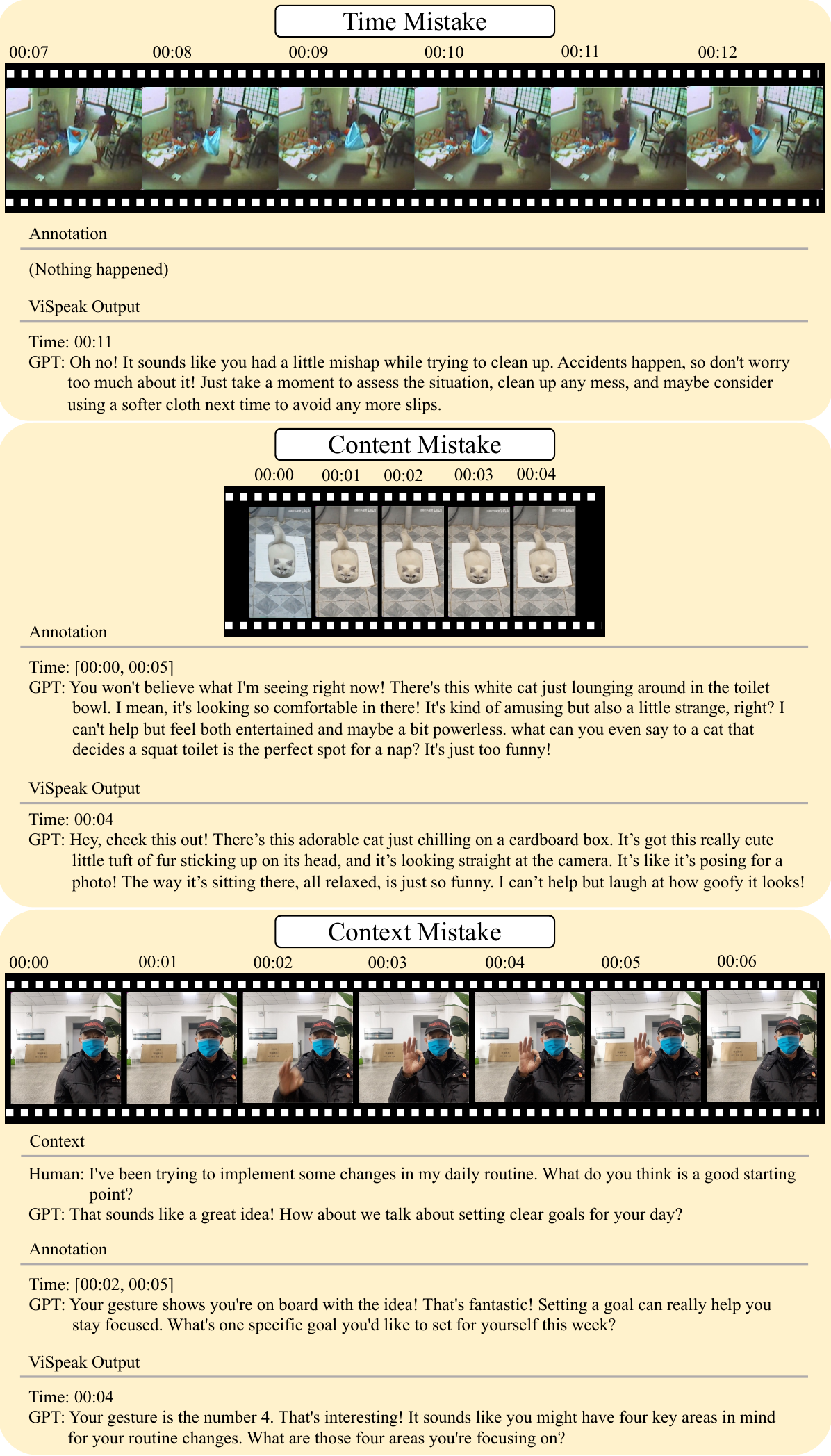}
  \caption{Examples of failure cases. The `Time Mistake' denotes the model responds at an improper time. The `Content Mistake' denotes the model fails to understand the visual content in the video. The `Context Mistake' means the model is unaware of the context of the conversation.}
  \vspace{-0.6em}
  \label{fig:failure_case}
\end{figure*}

In \Cref{fig:failure_case}, we visualize some failure cases of ViSpeak and mainly summarize them into three parts. First, ViSpeak may respond at an improper time. In the first example, there is nothing special in the video but ViSpeak begins to speak at 00:11 with some hallucinations. And ViSpeak may also ignore some actions and events. Second, ViSpeak may not understand the visual content in the video. As shown in the second video, the cat is actually in a toilet but ViSpeak mistakenly recognizes the toilet as a box thus failing to get the actual humor. In addition, ViSpeak may also not be aware of the context of the conversation. In the third example, the agent has asked the user about the feeling, not the number. But ViSpeak mistakenly recognizes the gesture as ``number 4''. Improvements in the future could solve the problems above to get a more intelligent agent.

\section{Limitation}
1) Due to the difficulty of the task and resource constraint, the diversity and scale of ViSpeak-Instruct are now relatively smaller than other well-known instruction following datasets. Expanding dataset size, collecting more diverse videos, and enriching more valuable sub-tasks are left for future work. 2) Second, due to computation constraints, ViSpeak is only trained with a 6k context. We believe a longer context will enhance users' experience. And a memory mechanism can equip the model with the long-term streaming video understanding ability. 3) Further, since we divide an integral audio into multiple small segments, we find the Automatic Speech Recognition (ASR) ability of ViSpeak degrades a lot, getting only 18.4 WER on LibriSpeech~\cite{panayotov2015librispeech}. Training with more audio data can possibly mitigate the problem. But we find ViSpeak still achieves SOTA performance on Omni-Source Understanding tasks of StreamingBench.

\end{document}